%% file: main.tex
\pdfoutput=1
\documentclass{article}
\usepackage{iclr2027_conference,times}
\iclrfinalcopy

\usepackage[T1]{fontenc}
\usepackage[utf8]{inputenc}
\usepackage{setspace}
\usepackage{hyperref}
\usepackage{url}
\usepackage{xurl}
\usepackage{graphicx}
\usepackage{microtype}
\usepackage{amsmath}
\usepackage{amssymb}
\usepackage{amsfonts}
\usepackage{booktabs}
\usepackage{multirow}
\usepackage{makecell}
\usepackage{array}
\usepackage{longtable}
\usepackage{tabularx}
\usepackage{multicol}
\usepackage{placeins}
\usepackage{needspace}
\usepackage{wrapfig}
\usepackage{subcaption}
\usepackage{enumitem}
\usepackage{xspace}
\usepackage[table]{xcolor}
\usepackage[export]{adjustbox}
\usepackage{tikz}
\usetikzlibrary{tikzmark,calc}
\usepackage{tcolorbox}
\tcbuselibrary{breakable,skins}
\makeatletter
\patchcmd{\@maketitle}{\LARGE\sc \@title\par}{\centering\normalfont\LARGE\bfseries \@title\par}{}{\PackageError{webharbor}{Could not apply title formatting}{The ICLR title block must match the expected template.}}
\patchcmd{\thebibliography}{\clubpenalty4000}{\clubpenalty10000}{}{}
\patchcmd{\thebibliography}{\widowpenalty4000}{\widowpenalty10000}{}{}
\makeatother

\newcommand{\ours}{\scalebox{0.95}{{\sffamily WebHarbor}}\xspace}

\hypersetup{
  hidelinks,
  pdfauthor={Zhaoyang Wang, Qianhui Wu, Boyu Gou, Demi Wang, Jiacheng Zhang, Peichun Hua, Qize Li, Shi Qiu, Yifan Jiang, Yifan Zhang, Zexu Jin, Baolin Peng, Bolin Shen, Chetan Bansal, Dehan Kong, Deren Lei, Dongshen Peng, Fengqing Jiang, Hao Cheng, Haoqing Wang, Jiaming Shan, Jicheng Wang, Jingguang Zhou, Jixuan He, Kaide Zeng, Kangyu Zhu, Lama Moukheiber, Mufan Qiu, Peng Xia, Qihan Sun, Qingwei Lin, Qiushi Sun, Ruichen Zhang, Shuqing Luo, Si Qin, Sicheng Fan, Tao Xiong, Wenlin Yao, Xinyu Zhang, Xinyue Zeng, Xuanrui Li, Xuchao Zhang, Yian Wang, Yichao Liu, Yiming Liang, Yiyang Zhou, Yizhu Jiao, Yuetai Li, Yurun Chen, Yuxuan Zhang, Zheyuan Yang, Zhiyuan Yao, Zhongyang Li, Zihao Li, Zili Wang, Jianfeng Gao, Huaxiu Yao},
  pdftitle={WebHarbor: Docking Real Websites for Evolving GUI Agent Environments}
}

\title{\raisebox{-0.12em}{\includegraphics[height=1.12em]{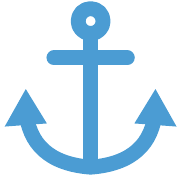}}\hspace{0.3em}WebHarbor: Docking Real Websites for\\Evolving GUI Agent Environments}

\author{\normalfont\parbox{\dimexpr\textwidth-2\tabcolsep\relax}{\centering\fontsize{10.8}{12.6}\selectfont\sffamily WebHarbor Team and Contributors\thanks{See full author list in Appendix~\ref{app:authors-contribution}.}}\\[\dimexpr3pt+\webharbortitlegap\relax]
\webharborcentred{\small\strut
\webharborlink{https://aiming-lab.github.io/WebHarbor/}{\faGlobe}{Project Page}\qquad
\webharborlink{https://huggingface.co/collections/ChilleD/webharbor}{\hflogo}{Data \& Models}\qquad
\webharborlink{https://github.com/aiming-lab/WebHarbor}{\faGithub}{Code}}}

\usepackage{fontawesome5} %
\definecolor{uncbolincreek}{HTML}{2C5080} %
\newcommand{\hflogo}{\raisebox{-0.15em}{\includegraphics[height=1.05em]{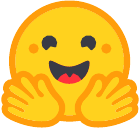}}}
\newcommand{\webharborlink}[3]{{\hypersetup{pdfborder={0 0 0}}%
  \href{#1}{\smash{#2}\ \textcolor{uncbolincreek}{#3}}}}
\newcommand{\webharborcentred}[1]{\hspace*{-\tabcolsep}\rlap{\makebox[\textwidth][c]{#1}}}
\newlength{\webharbortitlegap}
\makeatletter
\patchcmd{\@maketitle}{\hsize\textwidth}
  {\hsize\textwidth\vskip-\dimexpr\baselineskip+3pt\relax\vskip-\parskip}{}
  {\PackageWarningNoLine{webharbor}{Could not raise the title block}}
\patchcmd{\@maketitle}{\rule{\z@}{24pt}\@author}{\rule{\z@}{\dimexpr24pt-\webharbortitlegap\relax}\@author}{}
  {\PackageWarningNoLine{webharbor}{Could not tighten the title-author gap}}
\makeatother

\definecolor{unccloud}{HTML}{EDF5FB}
\newcommand{\webharborlogo}[1]{\raisebox{\dimexpr 8pt-0.5\height\relax}{#1}}
\fancypagestyle{webharborfirstpage}{%
  \fancyhf{}%
  \fancyhead[R]{\smash{%
    \webharborlogo{\includegraphics[height=15.5pt]{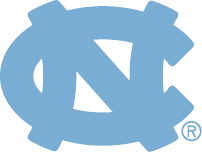}}\quad
    \webharborlogo{\includegraphics[height=16pt]{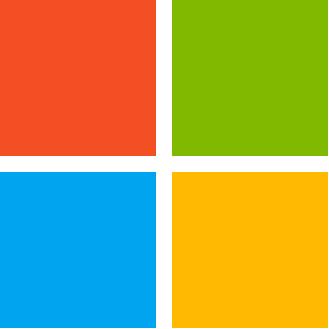}}}}%
  \fancyfoot[C]{\thepage}%
  \renewcommand{\headrulewidth}{0pt}%
}

\newcommand{\webharborabstractpanel}{%
  \tikz[remember picture,overlay]{%
    \iftikzmark{webharbor-abs-top}{\iftikzmark{webharbor-abs-bot}{%
      \coordinate (t) at (pic cs:webharbor-abs-top);%
      \coordinate (b) at (pic cs:webharbor-abs-bot);%
      \fill[unccloud,rounded corners=2mm]
        ($(t)+(0pt,14pt)$) rectangle ($(t|-b)+(\linewidth,-8pt)$);%
    }{}}{}}}
\renewenvironment{abstract}
  {\par\noindent
   \hbox to\linewidth{\tikzmark{webharbor-abs-top}\webharborabstractpanel\hss{\large\sc Abstract}\hss}%
   \par\begin{quote}}
  {\ifhmode\unskip\tikzmark{webharbor-abs-bot}\fi\par\end{quote}\par}

\begin{document}
\maketitle
\thispagestyle{webharborfirstpage}

\begin{abstract}
  GUI web agents require realistic and interactive environments that remain stable during both evaluation and training.
  Live website evaluation suffers from changing content, access restrictions, and limited coverage of authenticated workflows, while existing synthetic environments for training often lack functional depth and visual fidelity.
  We propose \ours, a suite of 153 executable and dockerized website environments that faithfully reconstruct the visual interfaces, content, and functional workflows of real-world web services under deterministic control.
  We build \ours through task-driven human-agent collaboration, where environments evolve as new tasks expand their functional scope, coding agents implement the required changes, and human contributors audit visual fidelity, functional correctness, and task validity.
  The resulting suite comprises 3,518 tasks spanning multi-step and post-authentication workflows, where existing works often fail to cover.
  Beyond reliable evaluation, \ours provides high-fidelity synthetic web environments for cost-efficient reinforcement learning (RL).
  We demonstrate that a Qwen 4B-level GUI agent trained with RL in \ours generalizes to unseen tasks and performs competitively with leading open-source GUI web agents across four benchmarks.
  The code and data are publicly available at \url{https://github.com/aiming-lab/WebHarbor}.
\end{abstract}

\section{Introduction}
\label{sec:introduction}

Web agents powered by large language models (LLMs) with multimodal capability interpret rendered webpages and complete tasks through browser interactions~\citep{deng2023mind2web,he2024webvoyager,zheng2024gpt4vision,qin2025uitars}.
Their progress depends on environments that support reliable evaluation and training.
Evaluation requires stable and reproducible outcomes, while reinforcement learning (RL) additionally requires executable state transitions and repeatable resets~\citep{qi2024webrl,bai2024digirl,wei2025webagentr1,lai2025computerrl}.
Live websites offer realistic interfaces, but changing content, bot detection, and regional restrictions undermine experimental control.
Moreover, live-web benchmarks rarely cover authenticated workflows, despite their importance to many everyday user requests~\citep{he2024webvoyager,xue2025illusion}.
In our experiments, about 38\% of Online-Mind2Web tasks were blocked by bot detection, and 22\% of WebVoyager tasks were outdated due to website changes.
These failures introduce uncontrolled variation into evaluation and prevent agents from learning the many workflows hidden behind authentication.

\input{figures/cmp_env_case_study.tex}
\input{tables/teaser_cmp.tex}

An alternative should combine realistic visual observations, functional interactions, and persistent authenticated state with offline execution and reliable resets.
Achieving this balance is challenging: commercial websites encompass extensive content and functionality, making complete reconstruction impractical, while visually similar replicas may lack correct search, account management, booking, or transaction logic.
Even apparently functional environments may expose shortcuts, omit plausible alternative actions, or leak answers through their interfaces, undermining task validity.
The goal is to match both what agents observe and how websites respond to their actions, while ensuring reproducibility and practical utility for both evaluation and RL.

Human crafted benchmarks such as WebArena and VisualWebArena offer reproducible applications but cover very few websites~\citep{zhou2023webarena,koh2024visualwebarena}.
Manually reconstructing real websites can improve fidelity, but requires substantial effort and remains difficult to scale.
Recent approaches automatically synthesize diverse functional websites~\citep{zhou2026webarenainfinity,zhang2026infiniteweb,wu2026autowebworld,cao2026guigenesis,wang2026cuagym}.
However, they typically emphasize newly generated worlds over task-grounded reconstructions of real websites, or offer limited multimodal content and post-authentication functionality.
Table~\ref{tab:teaser-comparison} summarizes this coverage gap: existing released environments do not jointly provide broad website coverage, offline execution, image-rich content, and login-based workflows.
The case study in Figure~\ref{fig:cmp-env-case-study} makes the visual-fidelity limitation concrete: the compared generated environments either omit the target website or simplify its layout, merchandising hierarchy, and product-specific imagery.

To address this gap, we introduce \ours, a suite of 153 executable website environments drawn from the website spaces covered by WebVoyager and Online-Mind2Web.
\ours uses task-conditioned construction: target tasks specify the pages, data, and interaction workflows that each environment must support, providing a tractable scope for reconstructing large web content.
Coding agents then implement the frontend, backend, database, and visual content through a standardized set of environment-construction skills.
Human contributors inspect the rendered websites, execute the target workflows, and revise failures in visual fidelity, functional correctness, data realism, and task validity, where fix implementations are done by coding agents.
This human-agent process produces Dockerized environments that run without access to the target services, reset to controlled initial states, and support both public information-seeking tasks and in-depth workflows behind authentication.
The final resource contains 3,518 tasks covering information retrieval, account management, bookings, and transactions, and can be used as a stable benchmark or as an executable training environment.

In summary, our contributions are threefold.
\begin{enumerate}[label=(\arabic*), leftmargin=0.6cm, nosep]
  \item We introduce \ours, a suite of 153 controllable web environments and 3,518 tasks with authenticated workflows and persistent state changes largely absent from live-web benchmarks.
  \item We develop a human-agent collaborative construction pipeline that combines task-conditioned website implementation by coding agents with iterative human auditing of visual fidelity, functional correctness, and task validity until approval.
  \item Experiments across three web agent benchmarks demonstrate that a Qwen 4B agent trained with RL in \ours generalizes to unseen tasks and achieves performance competitive with leading open-source web agents on live web benchmarks.
\end{enumerate}

\section{Related Work}
\label{sec:related}

\paragraph{GUI Web Agents}
Web agents have evolved from systems that operate on HTML or accessibility representations~\citep{zheng2024gpt4vision,he2024webvoyager} to multimodal agents that integrate webpage understanding, reasoning and planning, action grounding, and sequential decision-making to execute user instructions~\citep{gur2023realworld,koh2024tree,putta2024agent}. While prior work has largely focused on learning from demonstrations~\citep{hong2023cogagent,cheng2024seeclick,xu2024aguvis,qin2025uitars,wang2025opencua}, recent approaches such as WebRL, WebAgent-R1, and OpenWebRL, further optimize agent policies through reinforcement learning based on environment interactions~\citep{qi2024webrl,wei2025webagentr1,yang2026openwebrl}. These approaches motivate the need for environments that provide realistic visual observations, executable state transitions, and reproducible initial conditions. \ours provides such an infrastructure for efficient training and reliable evaluation of GUI web agents across diverse website interfaces.

\paragraph{Web Agent Benchmarks}
Existing benchmarks trade off website coverage, interaction depth, and experimental control.
MiniWoB++ and WebShop provide controlled settings for browser interaction and shopping~\citep{liu2018reinforcement,yao2022webshop}, while Mind2Web expands website coverage but relies on cached snapshots that limit interactive exploration~\citep{deng2023mind2web}.
Live-web benchmarks such as WebVoyager, Online-Mind2Web, and DeepShop offer realistic interactions across diverse websites~\citep{he2024webvoyager,xue2025illusion,lyu2025deepshop}.
Online-Mind2Web, even with periodic task updates, however, it remains vulnerable to website changes, bot detection, CAPTCHA, and network or geographic restrictions, while offering limited coverage of authenticated and state-changing workflows.
In contrast, self-hosted benchmarks such as WebArena, VisualWebArena, and WorkArena support complex interactions under controlled settings~\citep{zhou2023webarena,koh2024visualwebarena,drouin2024workarena}.
However, their limited website coverage, heavyweight deployments, and costly environment resets complicate repeated evaluation and online RL training.
REAL provides deterministic replicas of 11 real websites with persistent state changes, but starts from pre-authenticated states~\citep{garg2025real}.
\ours bridges these trade-offs with 153 image-rich and executable websites and 3,518 tasks, covering both information seeking and authenticated and state-changing workflows.
Its lightweight SQLite-backed architecture enables fast resets, facilitating reproducible evaluation and scalable online RL training across many websites.

\paragraph{Synthetic Web Environments}
Recent work scales web agent training through environment and task synthesis.
WebArena-Infinity and InfiniteWeb synthesize functional browser environments from documentation, specifications, and visual references~\citep{zhou2026webarenainfinity,zhang2026infiniteweb}.
AutoWebWorld, CUA-Gym, and WebForge further explore verifiable state transitions, task generation, and difficulty control~\citep{wu2026autowebworld,wang2026cuagym,yuan2026webforge}.
GUI-GENESIS and VeriEnv reconstruct real applications as executable environments for agent training~\citep{cao2026guigenesis,chae2026safe}.
However, synthesis pipeline with coding agent alone does not guarantee visual fidelity and realistic workflow depth.
In contrast, \ours emphasizes task-grounded reconstruction of real websites with human-validated quality.
We cover all websites from WebVoyager and Online-Mind2Web, extending task coverage to complex authenticated and state-changing workflows.
These tasks guide coding agents in iteratively implementing and expanding website functionality, while human reviewers audit visual fidelity, functional correctness, and task validity.
This approach yields 153 image-rich environments that combine broad website coverage with human-audited quality, supporting both reliable evaluation and scalable RL training.

\begin{figure}[t]
  \centering
  \includegraphics[max width=\linewidth]{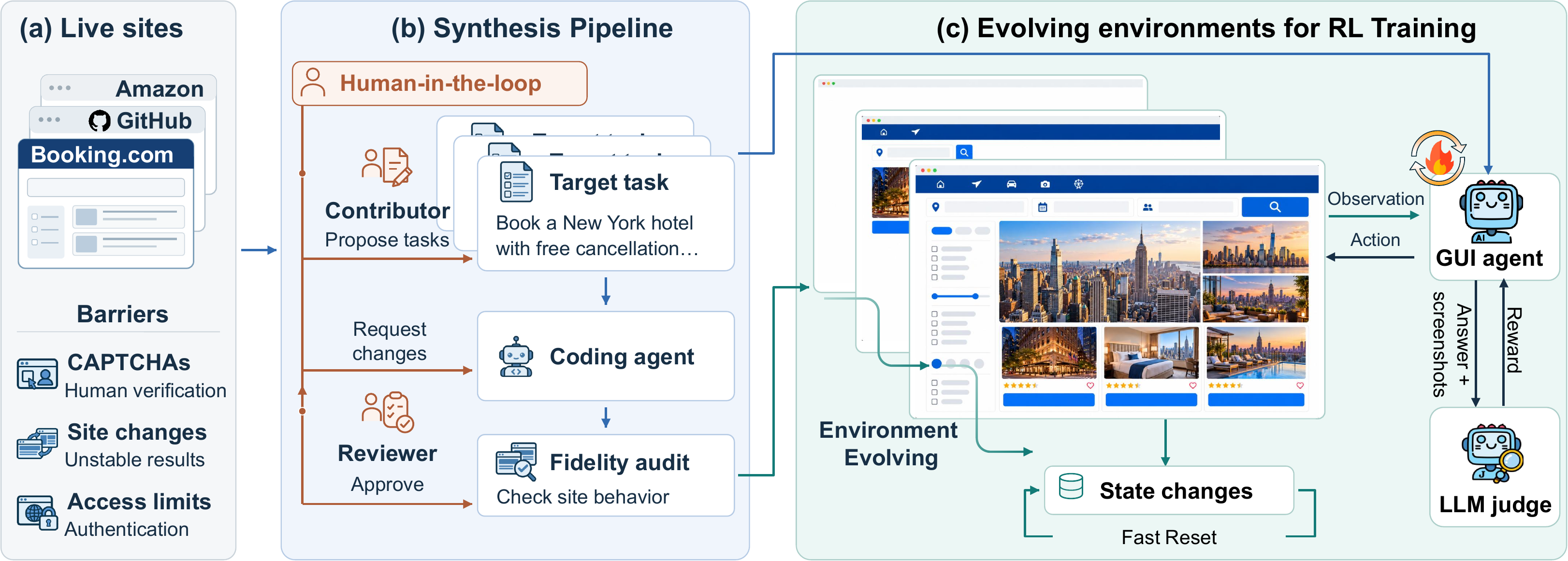}
  \caption{Overview of \ours. (a) Live websites provide target interfaces and workflows but introduce access barriers. (b) Target tasks scope full-stack reconstruction, then coding agents implement and evolve each environment, while human reviewers audit visual fidelity, functional correctness, and task validity until approval. (c) Dockerized environments support deterministic and fast state resets and evidence-grounded rewards for both evaluation and RL.}
  \label{fig:main}
\end{figure}

\section{WebHarbor}
\label{sec:method}

Given a target website and a set of relevant tasks, \ours reconstructs an executable local environment that preserves task-relevant interfaces, content, and state transitions.
We call this process \emph{docking}: replacing live dependencies with local assets, seeded state, and controlled resets while retaining recognizable visual and functional behavior.
Figure~\ref{fig:main} illustrates the iterative workflow.
Target tasks define the reconstruction scope, coding agents use reusable skills to implement the frontend, backend, database, and visual content, and human contributors audit the resulting environments.
Human feedback guides coding agents in repairing visual and functional defects.
As new tasks reveal missing functionality or shortcuts, the same process expands and refines the environments while maintaining deterministic execution and reliable resets.

\subsection{Task-driven Docking}
\label{sec:task-driven-docking}

Reconstructing an entire commercial website is impractical, given the significant scale of services such as Amazon's product catalog or YouTube's video library.
Instead, we adopt a task-driven reconstruction process that scopes each environment to the functionality and content required by its target tasks.
Each task specifies entry and terminal pages, observable entities and attributes, required browser actions, and persistent state changes.
These requirements define the task-relevant support surface: the pages, routes, image content, and data needed for realistic interactions.

We begin with WebVoyager's 15 websites and 643 tasks~\citep{he2024webvoyager}, whose live-web evaluation has become less reliable due to website changes.
We then extend coverage to all websites in Online-Mind2Web~\citep{xue2025illusion}, but exclude its original tasks, which primarily focus on information seeking rather than authenticated and state-changing workflows.
For these websites, human contributors identify core functional areas, and an LLM expands them into candidate workflows using its knowledge of common web services.
Human contributors then review and refine these tasks for functional coverage, feasibility, and interaction depth.
The resulting tasks cover account management, bookings, purchases, saved items, and other persistent operations.
Task-driven scoping also facilitates visual and functional fidelity by explicitly defining the observations and state transitions that must be reproduced.
Beyond task completion, environments preserve natural navigation paths and surrounding content, including plausible alternatives and near-miss items that prevent trivial shortcuts.
This approach bounds reconstruction effort while retaining the interaction complexity needed for evaluation and training.
The full list of 153 websites is provided in Tables~\ref{tab:full-envs-list} and~\ref{tab:full-envs-list-2} in the Appendix. Synthesis prompts are provided in Appendix~\ref{app:synthesis-prompts}.

\subsection{Skill-driven Synthesis and Evolution}
\label{sec:skill-driven-synthesis}

Task-driven docking defines what an environment should support, while skill-driven synthesis and evolution transform these requirements into an executable offline website.
For each website, the coding agent first explores the target service through a rendered browser, recording its task-relevant page hierarchy, navigation paths, URL patterns, interactive controls, and visual appearance.
The rendered DOM and resolved asset URLs provide references for reconstructing client-rendered pages and lazy-loaded content.
Guided by these observations and reusable construction skills (see Appendix~\ref{app:synthesis-prompts}), the coding agent builds a self-contained web application backed by SQLite and bundled with local visual assets.
These assets are either downloaded from the target website or replaced with visually similar alternatives obtained from public sources.
The initial reconstruction is then refined through repeated execution of target tasks.
For each task, the coding agent interacts with the rendered interface and identifies the first unsupported observation or state transition.
Missing pages prompt the creation of new routes and views, incomplete workflows require additional form handlers and persistent state updates, and insufficient content motivates new database records and task-relevant alternatives.
After each revision, the environment is reset to its seeded state, and the complete workflow is executed again.
This iterative process allows increasingly diverse and challenging tasks to expand and refine the environment until all target workflows can be completed end to end through the visible interface with the intended persistent state changes.

\subsection{Human-in-the-loop Iterative Refinement}
\label{sec:human-auditing}

Automated synthesis provides weak quality control for visual fidelity, long-horizon behavior, and task validity. Coding agents may complete target workflows while leaving surrounding interfaces visually inaccurate, functionally shallow, or artificially easy.
In our pilot study of 26 websites, 13 environments exhibited visual or asset defects (e.g., images mismatched with descriptions), and 14 had broken or shallow interactions (e.g., unresponsive buttons).
Such issues are often overlooked by coding agents during long-horizon environment construction but are readily identified by human reviewers.
Figure~\ref{fig:cmp-env-case-study} illustrates how fully automated reconstructions may capture coarse website layouts while simplifying functionality or omitting important visual content.

To improve the environment quality, we introduce human-in-the-loop auditing and iterative collaboration. Specifically, We organize human participation into two roles.
Contributors specify target websites and high-level functional requirements and guide coding agents during construction.
Designated reviewers then compare reconstructed environments against the target websites and execute each task end to end, checking task feasibility, intended interaction sequences, and visual and functional fidelity.
Reviewers translate identified defects into actionable requests for coding agents and re-evaluate affected workflows until approval.
On average, each environment undergoes 2--3 review rounds, each taking 1--2 hours.
Reviewers also develop task-specific verifiers and judging rubrics that define the grading criteria for LLM-as-a-Judge~\citep{zheng2023judging} or agent-as-a-Judge~\citep{zhuge2024agent}.
All contributors and reviewers are community volunteers included in the author list.

\subsection{Environment Suite}
\label{sec:suite-evaluation}
\input{figures/split_and_coverage.tex}
\input{tables/contribution_stat.tex}

Through approximately five months of community collaboration involving 57 contributors, iterative construction and auditing yielded 153 executable website environments with 3,518 tasks.
These environments support reliable resets, image-rich observations, and complex authenticated workflows for RL training and evaluation.
We construct the training set by holding out all 15 WebVoyager websites, 15 additional websites, and 20\% of tasks from each of the remaining 123 websites, yielding 2,288 training tasks.
The remaining tasks form the test candidate pool.
To ensure the reliability of test set, we execute these candidates using five models: GPT-4.1~\citep{gpt41}, GPT-5.5~\citep{gpt55}, GPT-5.6-Sol~\citep{gpt56}, GPT-6-Astra~\citep{gpt6}, and Qwen3.6-35B-A3B~\citep{qwen36_35b_a3b} across sizes and model capabilities.
We retain only tasks for which scripted verifiers and an OpenCode agent powered by GPT-5.6-Sol~\citep{opencode2025} agree on most executions.
To promote diversity, we further limit each website to two test tasks, producing a final test set of 306 tasks.
Figure~\ref{fig:train-split-exclusion} summarizes this final split.

\subsection{Dataset Analysis}
\input{tables/basic_stat_webharbor_transform.tex}
\input{tables/mirror_webvoyager_study.tex}
We analyze \ours along four dimensions.
(1) \textbf{Effective community collaboration}: As shown in Table~\ref{tab:community-collaboration}, 57 members across 26 affiliations collaboratively developed 153 website environments in five months. 
Substantial code contributions, cross-role participation, and iterative reviews demonstrate the feasibility of scaling environment construction through coordinated community effort.
(2) \textbf{Broad and deep task coverage}: Figure~\ref{fig:task-semantic-diversity} demonstrates that \ours achieves both broad semantic diversity and substantial interaction depth, rather than relying on repetitive information-seeking tasks.
Under a size-controlled comparison, \ours exhibits the lowest median nearest-neighbor similarity (0.58 versus 0.66--0.76), with its least-similar half covering approximately 91\% of websites.
The dashed cross-set curve shows a median nearest-task similarity of only 0.39 to Online-Mind2Web, indicating minimal task overlap and mitigating the risk of data leakage.
Moreover, Table~\ref{tab:method-task-coverage} presents that the suite includes 448 transaction/booking tasks, with 31.8\% requiring authentication, 35.6\% modifying persistent state, and 53.7\% involving cross-page reasoning, which remain largely absent from existing live-web and synthetic benchmarks.
(3) \textbf{Rich environment implementations}:
Table~\ref{tab:environment-suite-comparison} compares \ours to existing synthetic environment suites. \ours leads the compared suites in total and per-environment pages, images, asset size, and source lines, with an average of 37.3 pages, 1,319.9 images, 104.8 MiB of image assets, and 6,610.6 source lines of code per environment.
\input{figures/task_diversity.tex}
(4) \textbf{Consistent benchmark rankings}: Table~\ref{tab:webvoyager-ranking-consistency} shows that four agents achieve 4.4--18.8 points higher scores on \ours than on live WebVoyager,  reflecting reduced interference from content drift and access restrictions.
Crucially, all six pairwise agent rankings are preserved, with a Spearman rank correlation of 1.0 and a Pearson score correlation of 0.8, demonstrating strong consistency in relative agent performance across the two benchmarks.
These findings highlight the feasibility of community-driven construction, the diversity and depth of the resulting environments, and their value for both reliable evaluation and scalable RL training.
\WFclear

\section{Experiments}
\label{sec:experiments}
\input{tables/single_column_main_result1.tex}

\subsection{Experiment Setup}
\label{sec:exp_setup}

\paragraph{RL training \& Baseline}
We use OpenWebRL-4B~\citep{yang2026openwebrl} as our main baseline because it is a SOTA open-source agent trained through online RL on live websites. 
The synthetic environment suites WebArena-Infinity, AutoWebWorld, and CUA-Gym do not release trained checkpoints.
OpenWebRL is trained on 2,102 WebGym tasks across 704 live websites~\citep{bai2026webgym}. Our mirror-trained policy starts from the same SFT checkpoint, \texttt{OpenWebRL-4B-SFT}, and reuses the same multimodal multi-turn GRPO (MM-GRPO) objective and ReAct-style GUI harness.
At each turn, the policy observes the current state screenshot, textual action feedback, and full interaction history, and acts through browser tools (e.g., \textit{MOUSE\_CLICK}).
For \ours training, each instance runs in an isolated and deterministically seeded sandbox through Orchard interface~\citep{peng2026orchard}.
Table~\ref{tab:fast-reset-stat} measured latencies show that our environment supports efficient stateful episode turnover and timely verification suitable for large-scale RL training.
To keep the training framework and agent configuration consistent, we retain OpenWebRL's format-gated GPT-4.1 final-answer reward.
We isolate the environments and tasks as the primary difference, investigating whether RL in offline synthetic environments can match RL on live websites. Appendix~\ref{app:rl-training-implementation} gives more details.

\input{tables/fast_reset_stat.tex}
\paragraph{Benchmarks}
We evaluate generalization on four benchmarks. Online-Mind2Web contains 300 realistic tasks across 100+ live websites and provides broad coverage of sites and user domains~\citep{xue2025illusion}. WebVoyager comprises 643 tasks on 15 popular live websites~\citep{he2024webvoyager}.  WebArena contains complex tasks in reproducible self-hosted websites~\citep{zhou2023webarena}, where we use a subset of 226 tasks adopted by \citet{sun-etal-2025-os} and \citet{wang-etal-2026-synthagent}.  
To test authenticated workflows, WebArena-NL disables automatic login and provides the account credentials in the task description.
We also evaluate on the \ours test set, which is free from live-web disruptions and covers diverse websites.

\paragraph{Evaluation Protocol}
Following the evaluation protocol of OpenWebRL, we use a $1280\!\times\!1000$ viewport and decode with temperature 0.6, top-$p$ 0.95, and top-$k$ 20. Each task receives a fresh browser context and a 30-step budget except for WebArena, which uses a 50-step budget. 
Online-Mind2Web uses an o4-mini AgentTrek judge~\citep{xu2025agenttrekagenttrajectorysynthesis,xue2025illusion}. WebVoyager and DeepShop use benchmark-specific GPT-4.1 judges checking both the screenshots and the final answers. WebArena uses official programmatic state validators.
For \ours benchmark, we use the verifier to inspect state changes and the final answer.

\subsection{Main Results}
\input{tables/webharbor_test_res.tex}
\input{tables/single_column_main_result2.tex}

\paragraph{Unseen Benchmarks}
Tables~\ref{tab:single-column-main-result1} and~\ref{tab:single-column-main-result2} report four findings.
(1) \textbf{Consistent RL gains}. In the controlled 4B comparison, \ours-4B obtains the highest live-web average of 75.8, exceeding the shared SFT checkpoint by 22.2 points and the live-web-trained OpenWebRL-4B policy by 5.2 points. Its gains over OpenWebRL-4B hold across all four evaluation settings.
(2) \textbf{Open-source competitiveness}. \ours-4B achieves the strongest results among the listed open-source agents on both live-web benchmarks.
(3) \textbf{Complementary generalization}. All 15 WebVoyager websites are excluded from \ours training, the original Online-Mind2Web tasks are excluded from the \ours task suite, and WebArena uses stateful environments that were developed independently of \ours and do not appear in the training data. The results therefore demonstrate transfer to unseen websites, unseen task instances, and a distinct environment family.
(4) \textbf{Complex-workflow transfer}. Table~\ref{tab:single-column-main-result2} breaks WebArena down by site. \ours-4B leads on four of the five sites in both settings. When automatic login is disabled, its overall score drops by only 2.7 points, compared with 4.9 and 3.6 points for the OpenWebRL SFT and RL variants, and its lead widens on GitLab (30.4 vs.\ 19.6) and Map (25.0 vs.\ 15.6). Reddit is the exception, where all models stay below 12\% and \ours-4B solves only one of 26 tasks. The overall gain points to the authenticated workflows seen during training.
These findings suggest that training in \ours helps agents learn GUI policies that generalize well beyond the synthesized environments.

\paragraph{WebHarbor Benchmark}
Table~\ref{tab:webharbor-test-results} evaluates GPT-4.1, GPT-5.5, GPT-6-Astra, Qwen3.6-35B-A3B, and 4B baselines with a ReAct-style GUI harness and OpenCode equipped with GUI tools. High verifier--judge agreement supports test-set reliability. GPT-4.1 favors the GUI harness (26.4 vs. 1.0), whereas GPT-5.5 and Qwen3.6-35B-A3B benefit from OpenCode. \ours-4B outperforms all 4B baselines, reaching verifier and agent-judge pass rates of 30.0 and 41.5, respectively. GPT-6-Astra exceeds 95\% under both measures, indicating saturation for frontier models, yet the environments remain useful for training competitive small models, transferring gains to live websites, and synthesizing more complex computer-use tasks~\citep{wang2026agentworldmodelwild}.

\subsection{Analysis on RL Gains}
\label{sec:rl-generalization}
\input{figures/rl_study_by_task_type_float}
To locate where RL training helps, we classify each task with GPT-4.1 into capability groups. Figure~\ref{fig:rl-study-by-task-type} shows that \ours-4B improves over OpenWebRL-4B on every group in both benchmarks. Both overall differences are significant under paired tests ($p{<}0.01$). The largest gains fall on authentication (+13.3), multi-page navigation (+12.6), and stateful operations (+10.8) on \ours, and on transactions (+21.4) and stateful operations (+10.5) on WebArena. On Online-Mind2Web, 128 of the 139 tasks valid for all three agents are information retrieval. \ours-4B exceeds SFT checkpoint by 15.1 points on these tasks. The gains concentrate on authenticated and stateful workflows, which \ours offers and live web benchmarks largely lack in their released task collections.

\WFclear

\section{Conclusion}
\label{sec:conclusion}
We introduce \ours, a comprehensive suite of 153 executable website environments and 3,518 tasks that bring real-world web interactions into controllable offline settings.
Through task-driven docking and human-agent collaboration, these environments continuously evolve to support richer functionality while preserving visual fidelity and task validity.
The resulting suite offers broad website coverage, image-rich interfaces, and complex authenticated workflows, with lightweight state resets enabling reliable evaluation and efficient RL training.
Our held-out test set further provides a stable benchmark for assessing GUI web agent capabilities often overlooked by existing benchmarks.
Furthermore, we demonstrate that a Qwen 4B agent trained entirely in \ours generalizes effectively to unseen tasks and achieves competitive performance against leading open-source web agents across four benchmarks.
We believe \ours can facilitate reproducible evaluation, scalable RL training, and sustained community-driven evolution of GUI web agent environments.

\section*{Limitations}
As an environment suite for GUI web agents, \ours has several limitations.
\begin{enumerate}[label=(\arabic*)]
  \item \ours imitates target website rather than equaling it, and our claims are bounded by the proposed tasks and fidelity auditing by human reviewers. Client-heavy behavior (interactive maps, real-time tickers) is approximated or replaced with static surrogates.
  \item Despite judge consistency checks for the test set of \ours, scripted verifiers and model-based judges may overlook valid alternative solutions or subtle failures. This common limitation of web agent evaluation calls for more robust evaluation protocols in future work.
  \item To validate the training value of our synthesized offline environments, we mainly compare with OpenWebRL~\citep{yang2026openwebrl} which is a SOTA open-source agent trained through online RL on live websites. We do not systematically explore alternative supervised fine-tuning (SFT) data, general-purpose harnesses such as OpenCode~\citep{opencode2025} and CLI-based agents, or other RL algorithms. Due to computational constraints, we train only one 4B model and leave generalization across model families and scales to future work.
\end{enumerate}

\section*{Ethics Statement}

\ours is developed to support open research on GUI web agent evaluation and training.
Environment construction involves referencing real websites and obtaining visual assets from original services or publicly available alternatives.
We acknowledge that website designs, trademarks, and visual assets may be subject to third-party rights.
Our environments are intended for research and benchmarking rather than commercial replication of the original services.
Researchers reusing or redistributing these materials should respect applicable licenses and intellectual property rights.
By reconstructing websites as self-contained environments, \ours enables agents to practice complex interactions without repeatedly accessing or modifying live services.
Authenticated workflows, account management, bookings, and transactions operate within controlled local states, reducing risks associated with experiments on live websites.
However, synthetic environments cannot fully capture the diversity and dynamics of real websites, including evolving content, personalized interfaces, access restrictions, and external dependencies.
Strong performance in \ours does not guarantee equivalent performance or safe behavior on live websites.
Agents intended for real-world deployment require additional validation and appropriate safeguards.
We hope that accessible and reproducible environments will support more transparent and responsible research on GUI web agents.

\input{main.bbl}
\appendix
\clearpage
\input{appendix.tex}

\end{document}

%% file: figures/cmp_env_case_study.tex
\begin{figure}[t]
\centering
\includegraphics[width=0.9\linewidth]{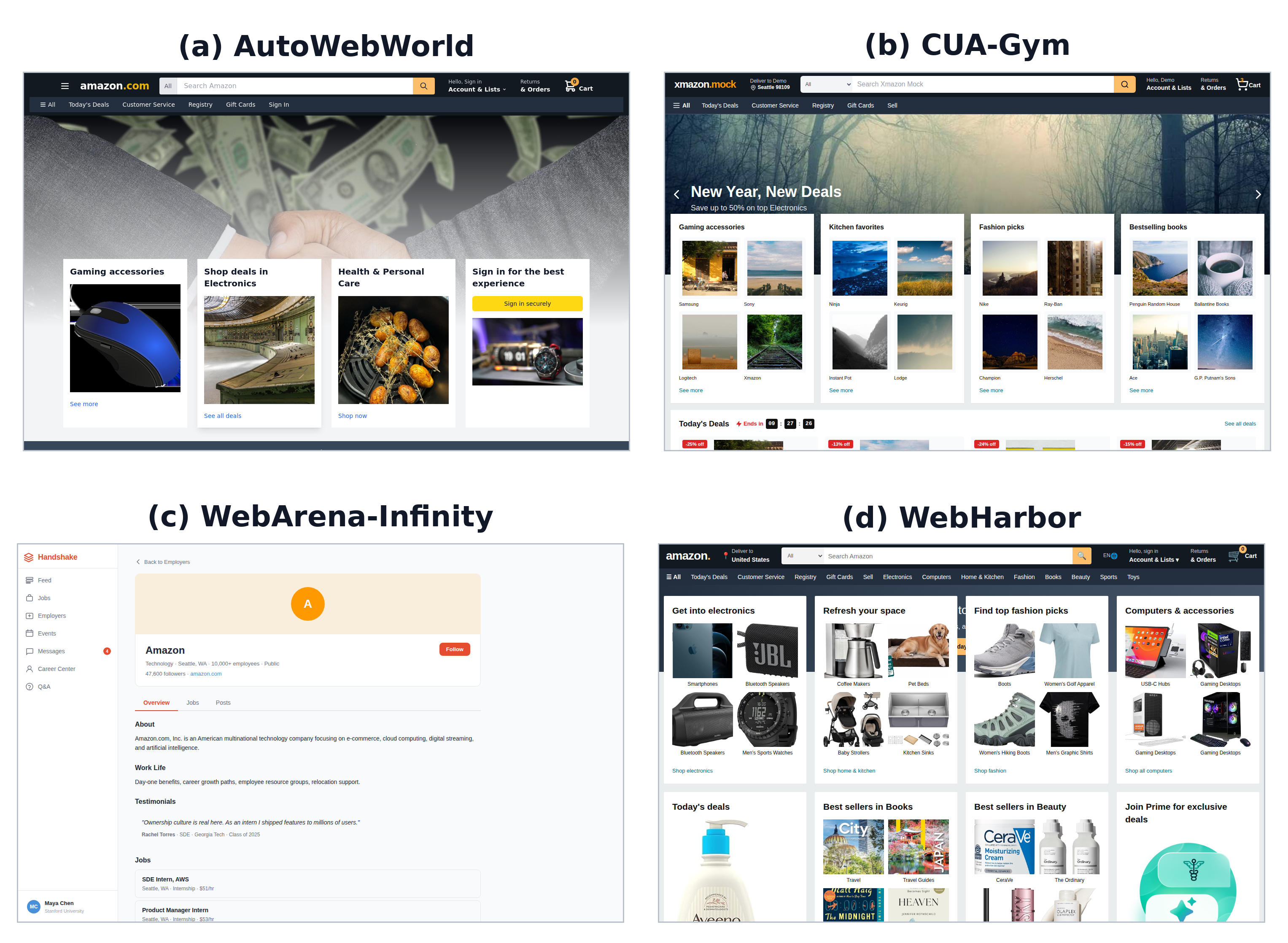}
\caption{Comparison on visual fidelity. (a) AutoWebWorld captures broad Amazon cues, but its hero and card layout depart from the origin. (b) CUA-Gym recovers the storefront structure while filling product slots with generic scenery images. (c) WebArena-Infinity provides no Amazon website, its closest page is a lightweight Handshake template without images. (d) WebHarbor combines target-like navigation, dense merchandising hierarchy, and product-specific imagery, yielding the most coherent and visually faithful environment. See more cases in Appendix~\ref{app:visual-case-studies}. \textbf{Zoom in for details}.}
\label{fig:cmp-env-case-study}
\end{figure}

%% file: tables/teaser_cmp.tex
\begin{table}[t]
\centering
\small
\begin{tabular}{@{}lccccc@{}}
\toprule
Method & Curation & \# Envs. & Offline & Images & Login \\
\midrule
\multicolumn{6}{@{}l}{\textit{Evaluation}} \\
WebVoyager~\citep{he2024webvoyager} & Human+LLM & 15 & $\times$ & \checkmark & $\times$ \\
WebArena~\citep{zhou2023webarena} & Human & 6 & \checkmark & \checkmark & \checkmark \\
Online-Mind2Web~\citep{xue2025illusion} & Human & 136 & $\times$ & \checkmark & $\times$ \\
DeepShop~\citep{lyu2025deepshop} & Agent & 1 & $\times$ & \checkmark & $\times$ \\
\midrule
\multicolumn{6}{@{}l}{\textit{Training}} \\
WebArena-Infinity~\citep{zhou2026webarenainfinity} & Agent & 13 & \checkmark & $\times$ & $\times$ \\
AutoWebWorld~\citep{wu2026autowebworld} & Agent & 29 & \checkmark & $\triangle$ & $\triangle$ \\
CUA-Gym~\citep{wang2026cuagym} & Agent & 98 & \checkmark & $\triangle$ & $\times$ \\
\midrule
\ours & Human+Agent & \textbf{153} & \checkmark & \checkmark & \checkmark \\
\bottomrule
\end{tabular}
\caption{Comparison of released web agent evaluation and training environments. Images denotes image-rich website content. $\triangle$ denotes generated images or partial coverage. Login requires the agent to authenticate through the interface. \# Envs. follow the actual releases.}
\label{tab:teaser-comparison}
\end{table}

%% file: figures/split_and_coverage.tex
\begin{figure}[t]
\centering
\small
\newsavebox{\webharborcoveragebox}
\sbox{\webharborcoveragebox}{%
\begin{tabular}{@{}lrrr@{}}
\toprule
Task capability & Tasks & Share (\%) & Sites \\
\midrule
Information/search & 2,282 & 64.9 & 151 \\
Account/profile & 161 & 4.6 & 98 \\
Content/communication & 100 & 2.8 & 46 \\
Transaction/booking & 448 & 12.7 & 95 \\
Settings/other state & 527 & 15.0 & 119 \\
\midrule
Authentication required & 1,117 & 31.8 & 135 \\
Persistent state change & 1,253 & 35.6 & 141 \\
Cross-page reasoning & 1,888 & 53.7 & 151 \\
\bottomrule
\end{tabular}}%
\begin{minipage}[t]{0.47\textwidth}
\vspace{0pt}\centering
\includegraphics[height=\dimexpr\ht\webharborcoveragebox+\dp\webharborcoveragebox\relax]{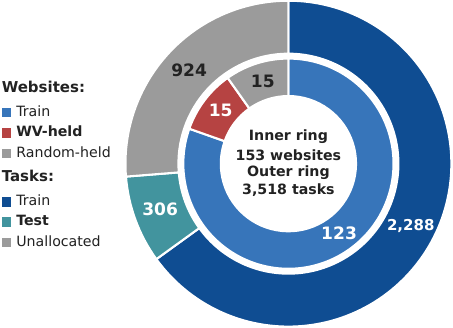}
\caption{Train/test split: 123 train and 30 held-out sites (inner), and 2,288 train, 306 test, and 924 unallocated tasks (outer).}
\label{fig:train-split-exclusion}
\end{minipage}\hfill
\begin{minipage}[t]{0.51\textwidth}
\vspace{0pt}\centering
\usebox{\webharborcoveragebox}
\captionof{table}{Task coverage by LLM-based classification. Rows 1--5 are exclusive categories, and rows 6--8 are non-exclusive properties.}
\label{tab:method-task-coverage}
\end{minipage}
\end{figure}

%% file: tables/contribution_stat.tex
\begin{table}[t]
    \centering
    \small
    \begin{tabular}{@{}lrrlrrr@{}}
    \toprule
    \multicolumn{7}{c}{Community composition} \\
    \cmidrule(lr){1-7}
    \multicolumn{7}{c}{%
        \setlength{\tabcolsep}{8pt}\begin{tabular}{@{}cccccc@{}}
        Members & Affiliations & Sites & Contrib. & Reviewers & Both roles \\
        57 & 26 & 153 & 33 & 29 & 22
        \end{tabular}%
    } \\
    \midrule
    \multicolumn{3}{c}{Project activity}
    & \multicolumn{4}{c}{Per-site collaboration} \\
    \cmidrule(lr){1-3}\cmidrule(lr){4-7}
    Activity & $N$ & Merged / median & Metric & Mean & Median & P90 \\
    \midrule
    GitHub PRs & 348 & 313 (89.9\%) & PRs & 3.18 & 3 & 5 \\
    HF asset PRs & 192 & 149 (77.6\%) & Participants & 3.33 & 2 & 5 \\
    PR comments & 316 & 132 words & Handoffs & 1.49 & 1 & 3 \\
    \bottomrule
    \end{tabular}
    \caption{Community collaboration in \ours. Some early-stage contributors are included in the total but may not be reflected in the role-specific counts. In Project activity, the third column reports the merged PR count (rate) for repository rows and median word count for PR comments. Handoffs count transitions between different authors within the same website.}
    \label{tab:community-collaboration}
\end{table}

%% file: tables/basic_stat_webharbor_transform.tex
\begin{table}[t]
\centering
\small
\begin{tabular}{@{}lrrrr@{}}
\toprule
\textbf{Metric} & WebArena-Infinity & AutoWebWorld & CUA-Gym & \textbf{\ours} \\
\midrule
\multicolumn{5}{@{}l}{\textit{Total}} \\
\midrule
Pages & 115 & 636 & 1,401 & 5,711 \\
Images & 0 & 2,303 & 56 & 201,946 \\
Assets (MiB) & 0.0 & 183.3 & 3.2 & 16,030.5 \\
Lines & 61,536 & 104,074 & 579,474 & 1,011,419 \\
\midrule
\multicolumn{5}{@{}l}{\textit{Mean (per environment)}} \\
\midrule
Pages & 8.8 & 21.9 & 14.3 & 37.3 \\
Images & 0.0 & 79.4 & 0.6 & 1,319.9 \\
Assets (MiB) & 0.00 & 6.3 & 0.03 & 104.8 \\
Lines & 4,733.5 & 3,588.8 & 5,913.0 & 6,610.6 \\
\bottomrule
\end{tabular}
\caption{Environment complexity and implementation breadth. Pages, Images, Assets, and Lines denote website routes, bundled image files, bundled image size (MiB), and source lines of code.}
\label{tab:environment-suite-comparison}
\end{table}

%% file: tables/mirror_webvoyager_study.tex
\begin{table}[t]
\centering
\small
\setlength{\tabcolsep}{3pt}
\begin{tabular*}{\linewidth}{@{\extracolsep{\fill}}lccccccc@{}}
\toprule
& \multicolumn{4}{c}{Model performance} & \multicolumn{3}{c}{Ranking consistency} \\
\cmidrule(lr){2-5}\cmidrule(l){6-8}
Benchmark & \makecell{Qwen3-VL-\\235B-A22B} & \makecell{Qwen3-VL-\\4B} & GPT-5.1 & \makecell{Orchard-4B\\\citep{peng2026orchard}} & \makecell{Pairwise\\orderings} & \makecell{Spearman\\$\rho$} & \makecell{Pearson\\$r$} \\
\midrule
WebVoyager & 66.4 & 52.6 & 55.9 & 74.1 & \multirow{3}{*}{\makecell{6/6\\(100\%)}} & \multirow{3}{*}{1.000} & \multirow{3}{*}{0.802} \\
\ours & 76.4 & 62.8 & 74.7 & 78.5 & & & \\
$\Delta$ & +10.0 & +10.2 & +18.8 & +4.4 & & & \\
\bottomrule
\end{tabular*}
\caption{Model performance and ranking consistency on WebVoyager and the same sites of \ours. Pairwise orderings counts model pairs whose order is preserved.}
\label{tab:webvoyager-ranking-consistency}
\end{table}

%% file: figures/task_diversity.tex
\begin{wrapfigure}[15]{r}{0.46\textwidth}
\vspace{-10pt}
\centering
\includegraphics[width=\linewidth]{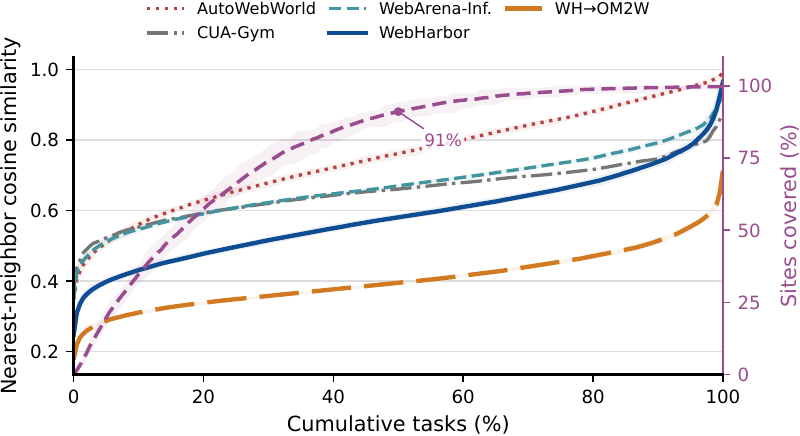}
\caption{Task semantic diversity: nearest-neighbor similarity within each suite and to Online-Mind2Web (left axis) and cumulative site coverage (right axis).}
\label{fig:task-semantic-diversity}
\vspace{-6pt}
\end{wrapfigure}

%% file: tables/single_column_main_result1.tex
\begin{table}[t]
\centering
\small
\begin{tabular}{@{}lccc@{}}
\toprule
\textbf{Model} & \multicolumn{1}{c}{\textbf{WebVoyager}} & \makecell[c]{\textbf{Online-}\\ \textbf{Mind2Web}} & \textbf{Avg.} \\
\midrule
\multicolumn{4}{@{}l}{\textit{Proprietary models}} \\
GPT-4o$^{*}$~\citep{openai2024gpt4ocard} & 65.1 & 34.6 & 49.9 \\
Gemini-3-flash$^{*}$~\citep{google2025gemini3flash} & 85.6 & 44.8 & 65.2 \\
GPT-5$^{*}$~\citep{openai2025gpt5} & \textbf{90.6} & 57.7 & \underline{74.2} \\
OpenAI CUA preview$^{*}$~\citep{openaicua} & 70.9 & 58.3 & 64.6 \\
Gemini CUA preview$^{*}$~\citep{google2025geminicua} & \underline{88.6} & 57.3 & 73.0 \\
\midrule
\multicolumn{4}{@{}l}{\textit{Open-source models}} \\
UI-TARS-1.5-7B$^{*}$~\citep{qin2025uitars} & 66.4 & 31.3 & 48.9 \\
GLM-4.1V-9B-Thinking$^{*}$~\citep{vteam2026glm45vglm41vthinkingversatilemultimodal} & 66.8 & 33.9 & 50.4 \\
Fara-7B$^{*}$~\citep{awadallah2025fara7befficientagenticmodel} & 73.5 & 34.1 & 53.8 \\
MolmoWeb-4B$^{*}$~\citep{gupta2026molmowebopenvisualweb} & 75.2 & 31.3 & 53.3 \\
MolmoWeb-8B$^{*}$~\citep{gupta2026molmowebopenvisualweb} & 78.2 & 35.3 & 56.8 \\
\midrule
\multicolumn{4}{@{}l}{\textit{4B model comparison}} \\
Qwen3-VL-4B-Thinking (base)    & 52.6\rlap{$^{*}$} & 32.0\rlap{$^{*}$} & 42.3 \\
OpenWebRL-4B-SFT               & 60.2\rlap{$^{*}$} & 47.0\rlap{$^{*}$} & 53.6 \\
OpenWebRL-4B (RL)              & 74.1\rlap{$^{*}$} & \underline{67.0}\rlap{$^{*}$} & 70.6 \\
\midrule
\textbf{\ours-4B}             & 81.2 & \textbf{70.4} & \textbf{75.8} \\
\bottomrule
\end{tabular}
\caption{Performance on WebVoyager and Online-Mind2Web, excluding invalid trials (e.g., reCAPTCHA). $^{*}$ marks results reported by OpenWebRL~\citep{yang2026openwebrl}. Bold and underlined values are the best and second-best results in each column.}
\label{tab:single-column-main-result1}
\end{table}

%% file: tables/fast_reset_stat.tex
\begin{wraptable}{r}{0pt}
\centering
\small
\begin{tabular}{@{}lrrr@{}}
\toprule
Metric & Mean (s) & Median (s) & P90 (s) \\
\midrule
Initialization & 7.43 & 6.75 & 9.37 \\
Reset & 6.68 & 6.00 & 9.00 \\
Verifier & 4.04 & 4.00 & 7.00 \\
\bottomrule
\end{tabular}
\caption{Latency of \ours operations including overhead of data transfer and Orchard management.}
\label{tab:fast-reset-stat}
\end{wraptable}

%% file: tables/webharbor_test_res.tex
\begin{table}[t]
\centering
\small
\begin{tabular}{@{}llccc@{}}
\toprule
\textbf{Model} & \textbf{Harness} & \textbf{Verifier} & \textbf{Agent-Judge} & \textbf{Consistency} \\
\midrule
\addlinespace
\multirow{2}{*}{GPT-4.1}
  & GUI & 26.4 & 37.9 & 93.5 \\
  & OpenCode    & 1.0  & 1.3  & 99.6 \\
\addlinespace
\multirow{2}{*}{GPT-5.5}
  & GUI & 26.9 & 27.4 & 98.6 \\
  & OpenCode    & 79.7 & 90.7 & 90.2 \\
\addlinespace
\multirow{2}{*}{GPT-6-Astra}
  & GUI & 86.7 & 95.4 & 90.3 \\
  & OpenCode    & 83.0 & 95.7 & 88.1 \\
\addlinespace
\multirow{2}{*}{Qwen3.6-35B-A3B}
  & GUI & 36.8 & 43.4 & 92.5 \\
  & OpenCode    & 57.4 & 66.0 & 88.7 \\
\midrule
\multicolumn{5}{@{}l}{\textit{4B web agents}} \\
Base & GUI & 19.7 & 24.5 & 94.9 \\
OpenWebRL-4B-SFT      & GUI & 29.1 & 35.5 & 90.6 \\
OpenWebRL-4B          & GUI & 19.7 & 25.7 & 91.6 \\
\midrule
\textbf{\ours-4B} & GUI & 30.0 & 41.5 & 87.0 \\
\bottomrule
\end{tabular}
\caption{Results on the 306-task \ours test set. Verifier and Agent-Judge report pass rates under different judge methods, and Cons. reports their judge agreement on trials with both verdicts.}
\label{tab:webharbor-test-results}
\end{table}

%% file: tables/single_column_main_result2.tex
\begin{table}[t]
\centering
\small
\begin{tabular*}{\linewidth}{@{\extracolsep{\fill}}lccccc|l@{}}
\toprule
\textbf{Model} & \textbf{Shopping} & \textbf{Admin} & \textbf{Reddit} & \textbf{GitLab} & \textbf{Map} & \textbf{Overall} \\
\midrule
\multicolumn{7}{@{}l}{\textit{WebArena (auto-login)}} \\
Qwen3-VL-4B-Thinking & 25.5 & 15.8 & 15.4 & 14.3 & 12.5 & 17.3 \\
OpenWebRL-4B-SFT & \textbf{36.4} & \underline{24.6} & \underline{23.1} & \underline{23.2} & \underline{18.8} & \underline{26.1} \\
OpenWebRL-4B (RL) & 29.1 & 21.1 & 7.7 & \underline{23.2} & 9.4 & 20.4 \\
\midrule
\textbf{\ours-4B} & \underline{32.7} & \textbf{33.3} & \textbf{26.9} & \textbf{25.0} & \textbf{21.9} & \textbf{28.8} \\
\midrule
\multicolumn{7}{@{}l}{\textit{WebArena-NL (no auto-login)}} \\
Qwen3-VL-4B-Thinking & \underline{30.9} & 12.3 & \underline{7.7} & 8.9 & 12.5 & 15.5$_{-1.8}$ \\
OpenWebRL-4B-SFT & \underline{30.9} & \underline{21.1} & \textbf{11.5} & \underline{19.6} & \underline{15.6} & \underline{21.2}$_{-4.9}$ \\
OpenWebRL-4B (RL) & 23.6 & 17.5 & \underline{7.7} & 16.1 & 12.5 & 16.8$_{-3.6}$ \\
\midrule
\textbf{\ours-4B} & \textbf{34.6} & \textbf{24.6} & 3.9 & \textbf{30.4} & \textbf{25.0} & \textbf{26.1}$_{-2.7}$ \\
\bottomrule
\end{tabular*}
\caption{Success rates (\%) of 4B models on WebArena by site, with and without automatic login (NL gives the credentials in the task instead). Subscripts give the drop from WebArena to WebArena-NL. Bold and underlined values are the best and second-best results in each column of each setting.}
\label{tab:single-column-main-result2}
\end{table}

%% file: figures/rl_study_by_task_type_float.tex
\begin{wrapfigure}{r}{0.42\textwidth}
\centering
\includegraphics[width=\linewidth]{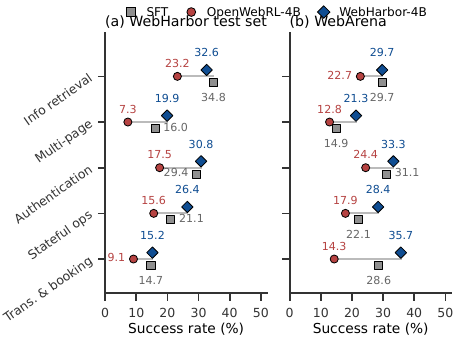}
\vspace{-14pt}
\caption{RL gains by task capability.}
\label{fig:rl-study-by-task-type}
\vspace{-6pt}
\end{wrapfigure}

%% file: appendix.tex
\input{authors_contributions.tex}

\section{Synthesis Prompts}
\label{app:synthesis-prompts}

\ours synthesizes each environment from observations of the target website and a task specification. Website observations define the page hierarchy, navigation, visual design, content structure, and interaction patterns. Tasks define the routes, records, account state, and state transitions required for evaluation. The synthesis process uses both sources to produce one executable environment.

Stage~1 reconstructs the interface, backend, data model, authentication flows, and initial seed state. Stage~2 defines tasks that cover the website's main functions and include multi-step workflows. Stage~3 executes every task and adds any missing routes, data, forms, or state transitions. Stage~4 removes answer leakage, adds near-miss alternatives, expands result sets, and checks cross-field consistency. Stage~5 finalizes the deterministic seed database and validates reset behavior.

Table~\ref{tab:prompt-clone-website} presents the Stage 1 prompt for reconstructing the target interface, backend, assets, authentication flows, and deterministic seed state. Table~\ref{tab:prompt-design-tasks} gives the Stage 2 prompt for defining executable task contracts with broad functional coverage, calibrated difficulty, and grounded answers. Table~\ref{tab:prompt-evolve-environment} details the Stage 3 execution loop that identifies missing routes, data, forms, and state transitions and then replays each repaired workflow. Table~\ref{tab:prompt-harden-environment} specifies the Stage 4 checks for answer leakage, near-miss distractors, catalog breadth, and cross-field consistency. Table~\ref{tab:prompt-seed-database} provides the Stage 5 prompt for idempotent seeding, benchmark account state, asset pinning, and reset verification. Each prompt also states its procedure, constraints, required checks, and output artifacts.

\input{prompts.tex}

\FloatBarrier

\section{Task Case Studies and Benchmark Comparison}
\label{app:task-case-study}
\input{tables/task_case_study.tex}

Tables~\ref{tab:task-case-study} and~\ref{tab:task-case-study-2} present ten authenticated tasks from Airbnb, Carvana, Compass, Thumbtack, and YouTube. Each task requires cross-page evidence, an authenticated action, and a persistent state change whose result the agent must then verify through the rendered interface.

The examples cover booking and reservation changes on Airbnb. They cover vehicle research, financing, and saved inventory on Carvana, as well as property collections and seller inquiries on Compass. The Thumbtack examples cover service requests, hiring, reviews, and messages. The YouTube examples cover subscriptions, comments, and Watch Later state. These workflows test whether an environment implements account-dependent operations.

\subsection{Task Examples Across Benchmarks}
\label{app:cross-benchmark-task-examples}

Tables~\ref{tab:cross-benchmark-task-examples} and~\ref{tab:cross-benchmark-task-examples-2} place two released tasks from each suite side by side. The task instructions show the operations evaluated within an episode and the account state available to the agent.

WebArena provides a strong controlled comparison. Its examples combine conditional reasoning, cross-site information gathering, and persistent writes. WebArena begins each episode from an authenticated browser state and evaluates subsequent account-dependent operations.

Online-Mind2Web and WebVoyager provide broad coverage of live websites, while their released tasks concentrate on interactions that are publicly accessible and require no authentication. All 300 Online-Mind2Web records are marked as not requiring login and include no stored session. Among WebVoyager's 643 released instructions, one explicitly asks the agent to sign in with credentials. The selected tasks illustrate their main scope: retrieval, filtering, and comparison over public content.

The synthesized suites also contain complex state-changing tasks. CUA-Gym supplies task-specific initial state, and WebArena-Infinity starts each generated application from seeded state. AutoWebWorld contains two explicit-login instructions among 4,322 unique released instructions. These systems demonstrate scalable generation of stateful interactions, but explicit authentication is rare.

\ours systematically includes login as part of task execution. The selected tasks require the agent to enter credentials, access account-specific state, complete multi-page actions, persist changes, and verify the final state from the interface. Under our fixed GPT-4.1 audit protocol, 1,117 of 3,518 \ours tasks are classified as authentication-dependent. Among them, 457 also require a state change and cross-page reasoning. Authentication is therefore evaluated as part of the task workflow.

\input{tables/cross_benchmark_task_examples.tex}
\FloatBarrier

\section{Environment Coverage and Visual Examples}
\label{app:visual-case-studies}
\input{figures/cmp_env_case_study_youtube.tex}

A GUI agent observes the rendered interface, including navigation, images, layout hierarchy, and metadata. Figure~\ref{fig:cmp-env-case-study-youtube} compares these properties across four YouTube-related environments. The comparison covers the browsing layout, video thumbnails, channel information, duration labels, and engagement metadata available to the agent.

Figures~\ref{fig:gallery_full} and~\ref{fig:gallery_full-2} show representative pages from 30 \ours environments across commerce, travel, media, entertainment, developer tools, and information services. The gallery records the variation in page structure, content density, imagery, navigation, and controls covered by the suite.

\input{figures/our_envs_gallery.tex}

\subsection{Environment Inventory}
\label{app:environment-coverage}

Tables~\ref{tab:full-envs-list} and~\ref{tab:full-envs-list-2} list all 153 \ours environments and their source-benchmark website coverage. Online-Mind2Web groups websites by domain and reports 136 websites. We separate services that share a domain into individual environments, yielding 147 environments from Online-Mind2Web. Adding the 15 WebVoyager environments, nine of which are shared, gives 153 in total.

\FloatBarrier
\input{tables/full_envs_list.tex}
\FloatBarrier

\section{RL Training Implementation Details}
\label{app:rl-training-implementation}

We use the established OpenWebRL framework~\citep{yang2026openwebrl} to evaluate \ours as an RL training substrate. The released live-web-trained reference policy and our mirror-trained policy use the same \texttt{OpenWebRL-4B-SFT} initialization, multimodal multi-turn GRPO objective, browser-agent interface, and GPT-4.1 reward definition. The \ours adaptation replaces the live task pool and browser backend with \ours tasks executed in deterministic mirror sandboxes. This design isolates the training signal of the mirrors from any change in policy optimization.

\subsection{Browser Agent Harness}
\label{app:rl-browser-harness}

At each turn, the policy receives the current screenshot, active URL, viewport and tab metadata, textual feedback from the previous action, and the complete reasoning and tool-call history. Only the current screenshot remains in the multimodal context. Textual history preserves earlier observations and action outcomes. Table~\ref{tab:rl-browser-action-space} lists the 13 browser tools shared by live-web and mirror rollouts. The harness may execute several tool calls from one response before returning the next observation.
\input{tables/rl_browser_action_space.tex}
\input{tables/rl_training_hyperparameters.tex}

\subsection{Task Verification and Agent Judge}
\label{app:verification-agent-judge}

Each episode executes one task in an isolated image for its website. At startup, the image copies the frozen seed into a fresh live database and stores a root-only initial snapshot before the agent begins. The harness records the action--observation trajectory, final answer, termination state, browser state, and timestamped screenshots. Root-owned evidence separately preserves the site-access log and the initial and post-episode database states. The verifier assembles these records after the agent phase so that grading uses the observed interaction and protected environment state.

The reviewer authors a task-specific verifier and a natural-language rubric as two forms of the grading contract. The verifier produces the official binary outcome. Depending on the task, it checks required answer facts, task-relevant visited routes, the relevant final browser state, and the exact database delta between the initial and final snapshots. Mutation checks require the requested additions, updates, or deletions and reject changes to unrelated tables, rows, or fields. Read-only tasks require state preservation. The rubric specifies the success conditions, acceptable semantic variants, required interaction or sequence evidence, and failure cases for semantic review. It is used only during grading and is never shown to the acting agent.

The deterministic verifier runs first. The Agent Judge then runs in the same sandbox before it is released and audits the completed trial. It receives the task instruction, reviewer-authored rubric, complete normalized trajectory, final answer, and screenshots. It may also inspect the verifier code and recorded verdict, pristine seed database, live final database, and site code. It returns \texttt{agent\_success}, \texttt{agent\_failed}, or \texttt{task\_error}, together with a confidence score, a concise explanation, verified hard facts, and a verifier-disagreement flag. The \texttt{task\_error} label is reserved for a concrete environment defect that makes the task impossible, never for an agent mistake. The Judge verdict is an audit signal and cannot alter the verifier result. A missing or malformed Judge output is recorded as a Judge error while the verifier result remains unchanged. The two signals therefore cover different requirements: the verifier provides a reproducible final-outcome check, while the Agent Judge can assess semantic equivalence and sequence-dependent rubric items from the trajectory. Both remain separate from the optimized OpenWebRL reward described below.

\subsection{Training Configuration}
\label{app:rl-training-configuration}

Training starts from the released OpenWebRL SFT checkpoint and applies MM-GRPO to groups of five trajectories per task. A trajectory-level reward is converted to a group-relative advantage for every assistant turn, and groups with identical rewards are removed by dynamic sampling. The reward is OpenWebRL's format gate followed by a GPT-4.1 success judgment over the final answer, action history, and recent screenshots. The task-specific \ours verifier serves only as a diagnostic metric. Table~\ref{tab:rl-training-hyperparameters} lists the full rollout, optimization, and reward settings.

%% file: authors_contributions.tex
\section{Authors \& Contribution}
\label{app:authors-contribution}

\begingroup
\small
\renewcommand{\thefootnote}{\fnsymbol{footnote}}
\newcommand{\whperson}[2]{\mbox{#1\textsuperscript{#2}}}
\begingroup
\onehalfspacing
\paragraph[Project leads]{Project leads\protect\footnote[2]{\label{fn:author-correspondence}\raggedright Correspondence to Zhaoyang Wang (\href{mailto:zhaoyang@cs.unc.edu}{\mbox{\fontsize{8.5}{10}\selectfont\urlstyle{same}\nolinkurl{zhaoyang@cs.unc.edu}}}) and Qianhui Wu (\href{mailto:qianhuiwu@microsoft.com}{\mbox{\fontsize{8.5}{10}\selectfont\urlstyle{same}\nolinkurl{qianhuiwu@microsoft.com}}}).\endgraf}}
\whperson{Zhaoyang Wang}{1}, \whperson{Qianhui Wu}{2}

\paragraph[Core contributors]{Core contributors\protect\footnote[1]{\label{fn:author-order}Authors in these two groups are listed alphabetically by first name.}}
\whperson{Boyu Gou}{3}, \whperson{Demi Wang}{4}, \whperson{Jiacheng Zhang}{5}, \whperson{Peichun Hua}{6}, \whperson{Qize Li}{7}, \whperson{Shi Qiu}{1}, \whperson{Yifan Jiang}{8}, \whperson{Yifan Zhang}{9}, \whperson{Zexu Jin}{7}

\paragraph[Contributors]{Contributors\textsuperscript{\hyperref[fn:author-order]{*}}}
\whperson{Baolin Peng}{2}, \whperson{Bolin Shen}{10}, \whperson{Chetan Bansal}{2}, \whperson{Dehan Kong}{7}, \whperson{Deren Lei}{11}, \whperson{Dongshen Peng}{12}, \whperson{Fengqing Jiang}{13}, \whperson{Hao Cheng}{2}, \whperson{Haoqing Wang}{14}, \whperson{Jiaming Shan}{15}, \whperson{Jicheng Wang}{16}, \whperson{Jingguang Zhou}{4}, \whperson{Jixuan He}{1}, \whperson{Kaide Zeng}{1}, \whperson{Kangyu Zhu}{1}, \whperson{Lama Moukheiber}{17}, \whperson{Mufan Qiu}{1}, \whperson{Peng Xia}{1}, \whperson{Qihan Sun}{14}, \whperson{Qingwei Lin}{2}, \whperson{Qiushi Sun}{18}, \whperson{Ruichen Zhang}{1}, \whperson{Shuqing Luo}{1}, \whperson{Si Qin}{2}, \whperson{Sicheng Fan}{19}, \whperson{Tao Xiong}{20}, \whperson{Wenlin Yao}{2}, \whperson{Xinyu Zhang}{6}, \whperson{Xinyue Zeng}{21}, \whperson{Xuanrui Li}{14}, \whperson{Xuchao Zhang}{2}, \whperson{Yian Wang}{22}, \whperson{Yichao Liu}{23}, \whperson{Yiming Liang}{21}, \whperson{Yiyang Zhou}{1}, \whperson{Yizhu Jiao}{24}, \whperson{Yuetai Li}{13}, \whperson{Yurun Chen}{20}, \whperson{Yuxuan Zhang}{25}, \whperson{Zheyuan Yang}{26}, \whperson{Zhiyuan Yao}{20}, \whperson{Zhongyang Li}{3}, \whperson{Zihao Li}{24}, \whperson{Zili Wang}{16}, \whperson{Jianfeng Gao}{2}, \whperson{Huaxiu Yao}{1}
\par
\endgroup

\subsection*{Affiliations}
\begin{multicols}{2}
\raggedcolumns
\raggedright
\hyphenpenalty=10000
\exhyphenpenalty=10000
\begin{enumerate}[label=\arabic*.,leftmargin=*,nosep]
\item University of North Carolina at Chapel Hill
\item Microsoft
\item The Ohio State University
\item Carnegie Mellon University
\item University of Michigan
\item The Chinese University of Hong Kong
\item WebAgentLab
\item University of Southern California
\item University of Delaware
\item Florida State University
\item Meta Superintelligence Labs
\item Stanford University
\item University of Washington
\item Queen Mary University of London
\item University of California, Santa Barbara
\item University of California, Davis
\item Georgia Institute of Technology
\item The University of Hong Kong
\item Fudan University
\item Zhejiang University
\item Virginia Tech
\item National University of Singapore
\item Nankai University
\item University of Illinois Urbana-Champaign
\item University of British Columbia
\item Tongji University
\end{enumerate}
\end{multicols}
\par
\endgroup

%% file: prompts.tex
\input{tables/prompts.tex}

%% file: tables/prompts.tex
\providecolor{PromptNavy}{HTML}{17324D}
\providecolor{PromptRule}{HTML}{7F91A3}

\newtcolorbox{skillprompt}[2]{enhanced, colback=white, colframe=PromptRule,
  boxrule=0.5pt, arc=1.5pt, left=5pt, right=5pt, top=3pt, bottom=4pt, toptitle=2pt, bottomtitle=2pt,
  colbacktitle=white, coltitle=PromptNavy, titlerule=0.4pt,
  fonttitle=\sffamily\bfseries, title={#1\hfill{\mdseries\footnotesize Skill: \texttt{#2}}},
  fontupper=\ttfamily\footnotesize\raggedright, before skip=8pt, after skip=6pt}

\input{tables/prompt_clone_website.tex}
\input{tables/prompt_design_tasks.tex}
\input{tables/prompt_evolve_environment.tex}
\input{tables/prompt_harden_environment.tex}
\input{tables/prompt_seed_database.tex}

%% file: tables/prompt_clone_website.tex
\begin{table}[tp]
\begin{skillprompt}{Stage 1: Clone the target website}{clone-website}
\textbf{Objective.} Build a new executable \ours mirror from the observed website, with browser-captured evidence, aligned real assets, working interactions, authenticated state, and deterministic seed data. Target site: \textless{}REAL\_URL\textgreater{}. Site slug: \textless{}SLUG\textgreater{}.\par
\textbf{Scaffold and register.} Run \texttt{./scripts/new\_site.py \textless{}SLUG\textgreater{}}. Keep the prescribed \texttt{sites/\textless{}SLUG\textgreater{}/} layout and register the site consistently in \texttt{websyn\_start.sh}, \texttt{control\_server.py}, and \texttt{Dockerfile}. Derive the port from the current registry. Do not leave reports, review notes, or one-off harvest scripts inside the site directory.\par
\textbf{Observe the live site.} Define a tracked snapshot contract with capture ID and time, locale, timezone, viewport, source URLs, requested and final states, and retained screenshot/DOM evidence. Drive a real Chromium browser with Playwright. Capture the post-hydration DOM, resolved image URLs, navigation hierarchy, URL patterns, listing and detail views, search, authentication, account pages, and state-changing forms. Do not use \texttt{requests} or \texttt{curl} as a substitute for rendered-page observation.\par
\textbf{Build the mirror.} Harvest real target images and brand assets; record stable entity keys, upstream URLs, local paths, and SHA-256 digests. Implement Flask and SQLAlchemy models, password-backed authentication, search, account state, entity detail pages, and required CRUD flows. Build Jinja2 templates that reproduce the target's hierarchy, typography, navigation, and responsive behavior. Every visible filter, tab, sort control, form, and persistent action must change meaningful state.\par
\textbf{Seed and verify.} Gate each seed function as a whole, create benchmark users and realistic histories, boot the site, and freeze \texttt{instance\_seed/\textless{}SLUG\textgreater{}.db}. Build the current image and test the mirror on isolated ports. Use Playwright to verify decoded images, entity-to-image alignment, core page types, narrow and desktop viewports, major interactions, saved-state persistence after restart, and byte-identical database restoration after reset. Stop only the test container owned by this run.\par
\textbf{Deliverables.} Return the registered site code, templates, real assets and provenance, stable seed database, retained visual evidence, completed checks, and any unresolved blocker. Do not proceed to task design until the mirror is reachable, non-empty, interactive, and resettable.
\end{skillprompt}
\caption{Skill prompt used in Stage 1 to construct an executable clone of the target website.}
\label{tab:prompt-clone-website}
\end{table}

%% file: tables/prompt_design_tasks.tex
\begin{table}[tp]
\begin{skillprompt}{Stage 2: Design task contracts}{design-tasks}
\textbf{Objective.} Create 15--20 benchmark tasks for \texttt{sites/\textless{}SLUG\textgreater{}/tasks.jsonl}. Treat every task as an executable contract between the natural-language instruction, the mirror's interface, its database state, and the verifier. Reuse applicable WebVoyager tasks verbatim, then add tasks grounded in the target site's real workflows.\par
\textbf{Schema.} Write one JSON object per line with \texttt{web\_name}, a unique \texttt{id} of the form \texttt{\textless{}web\_name\textgreater{}--\textless{}index\textgreater{}}, the user-facing \texttt{ques}, the local mirror URL in \texttt{web}, and the real \texttt{upstream\_url}. Keep instructions natural and include demo credentials directly when authentication is required. Do not place expected answers or grading hints in the task text.\par
\textbf{Functional coverage.} Cover the site's complete feature surface rather than repeating one search flow. Include search and filtering, browsing and detail inspection, account management, saved state, history, comparison, and the site's characteristic state-changing operations such as cart, checkout, booking, messaging, review, subscription, or collection management. Use several tasks from each major functional area.\par
\textbf{Difficulty.} Include easy tasks of two or three steps, medium tasks of four to six steps, and at least 3--5 hard tasks requiring five or more steps. Hard tasks should require cross-page evidence, visual comparison, constraint composition, arithmetic or ranking, or a persistent action followed by confirmation. Include at least two genuine disambiguation tasks for which the seeded account contains multiple plausible items and the agent must request clarification instead of guessing.\par
\textbf{Grounding.} Ensure the database contains a stable, page-specific answer for every non-disambiguation task. Natural searches must return at least six plausible results, including near misses. The answer must not be exposed in a result-card title, count label, snippet, sort order, or unique single result when the task is intended to require detail-page reasoning.\par
\textbf{Verification.} Hand-walk every task through the rendered mirror with Playwright. Confirm that all required routes and controls are reachable, the correct answer or state transition exists, authenticated actions persist after reload, and ambiguous tasks expose at least two valid choices. Record missing capabilities for Stage 3. Report task count, functional-area coverage, difficulty distribution, reused source tasks, and every task that still requires environment evolution.
\end{skillprompt}
\caption{Skill prompt used in Stage 2 to define executable task contracts for the cloned website.}
\label{tab:prompt-design-tasks}
\end{table}

%% file: tables/prompt_evolve_environment.tex
\begin{table}[tp]
\begin{skillprompt}{Stage 3: Task-driven environment evolution}{evolve-env}
\textbf{Objective.} Execute every task in \texttt{sites/\textless{}SLUG\textgreater{}/tasks.jsonl} and evolve the mirror until each workflow can be completed naturally through the rendered interface. Preserve the target site's interaction conventions and the byte-identical reset invariant while adding missing capabilities.\par
\textbf{Browser execution.} Use Playwright with a real Chromium browser for every search, navigation, form submission, login, and state verification. Save evidence at the important steps. Do not substitute \texttt{curl}, \texttt{requests}, Flask's test client, or static template inspection for browser interaction; those methods do not expose client-side validation, route guards, asynchronous redirects, hidden fields, or the DOM seen by the agent.\par
\textbf{Task loop.} For each task, begin from the deterministic seed, perform the instruction end to end, and identify every missing route, template, database record, form handler, validation rule, or state transition. Extend \texttt{app.py}, Jinja templates, and \texttt{seed\_data.py} only as required by consistent site behavior. Rebuild and replay the complete workflow after each change. Persistent actions must remain visible after navigation and reload, and unrelated account state must remain unchanged.\par
\textbf{Detect shortcuts.} Inspect the rendered search and listing pages for task-answer leaks, precomputed answers, unique obvious targets, or result ordering that reveals the solution. Exercise malformed and empty forms, partial queries, multiple detail pages, login persistence, and back-navigation. Replace exact-match-only search with scored matching where needed. Add realistic category-adjacent records when more than half of the returned items satisfy all task constraints.\par
\textbf{Data discipline.} Gate every seed function at the function level. Seed four benchmark accounts under \texttt{@test.com} with the documented shared password and useful pre-existing carts, bookmarks, orders, histories, or messages. After database changes, regenerate the seed database from a clean boot. Use paths derived from \texttt{BASE\_DIR}; do not hard-code machine-specific locations.\par
\textbf{Verification.} Build the current image, launch an isolated test container, reset the affected site, and compare runtime and seed database bytes. Also test restart separately: create persistent state through the UI, restart the site without resetting it, and confirm that the state survives. Re-run all tasks in Playwright and verify decoded images, functional controls, persisted state, and complete workflows.\par
\textbf{Deliverables.} Report the pass/fail result for every task, the capability added for each failure, retained screenshots, seed-byte comparison, and remaining hardening targets. Finish only when every task is executable and the mirror has no broken form, empty destination, placeholder page, or obvious answer leak.
\end{skillprompt}
\caption{Skill prompt used in Stage 3 to evolve the environment through complete task execution.}
\label{tab:prompt-evolve-environment}
\end{table}

%% file: tables/prompt_harden_environment.tex
\begin{table}[tp]
\begin{skillprompt}{Stage 4: Environment hardening}{harden-env}
\textbf{Objective.} Audit every executable task along four dimensions: answer de-leaking, near-miss distractors, catalog breadth, and cross-field consistency. The environment must preserve task feasibility while preventing an agent from succeeding through a visible answer, a single obvious result, or inconsistent data.\par
\textbf{A. De-leak answers.} Search with the task's natural query and inspect the rendered DOM and screenshot before opening any detail page. Remove constraint values from item names, answer text from snippets, precomputed comparisons, visible list counts, question-shaped headings, pinned answer callouts, and algorithm-revealing labels. Check all known leak patterns, including fuzzy search that floods an answer bucket, a sort order that places the target first, and a pre-curated lookup table that reproduces the task wording. Move required evidence into consistent specs, descriptions, histories, or detail views.\par
\textbf{B. Add near misses.} For each task, retain 1--3 full matches and add at least five plausible items that match the category but fail exactly one constraint, such as price, size, color, date, brand, location, availability, or required feature. Keep the full-match density at or below 50\%. The alternatives must look realistic and support meaningful comparison; do not add arbitrary noise.\par
\textbf{C. Broaden the catalog.} Require each natural query to return at least six items from multiple relevant subcategories. Expand thin category pages and search indexes with real-looking entities, useful metadata, and aligned images. Preserve the target site's hierarchy so broader coverage increases decision complexity without changing the intended answer.\par
\textbf{D. Enforce consistency.} Treat structured specifications as the source of truth. Whenever a field changes, regenerate every dependent description, feature bullet, tag, variant option, availability label, and displayed summary. Confirm that listing cards, detail pages, account records, and transaction history agree. Keep constraint values out of names unless the real site uses them as product identity.\par
\textbf{Per-task verification.} Use Playwright to run the natural query, capture the results, search the rendered DOM for the answer, count full matches and near misses, verify result diversity, click the correct detail page through the interface, and compare all related fields. Record pass/fail evidence for all four dimensions. Static template review and HTTP probes do not replace this check.\par
\textbf{Final invariants.} Rebuild after every data change, reset the site, and confirm byte equality between runtime and seed databases. Re-run authenticated and state-changing workflows, inspect desktop and narrow viewports, and verify that all required images decode. Finish with zero answer leaks, at least five near misses per task, broad search results, consistent fields, and unchanged task solvability.
\end{skillprompt}
\caption{Skill prompt used in Stage 4 to strengthen environment fidelity, consistency, and task difficulty.}
\label{tab:prompt-harden-environment}
\end{table}

%% file: tables/prompt_seed_database.tex
\begin{table}[tp]
\begin{skillprompt}{Stage 5: Finalize the seed database}{seed-database}
\textbf{Objective.} Finalize the deterministic seed database and the asset side of the mirror. The accepted runtime state must survive ordinary restart, return exactly after reset, support realistic search and authenticated histories, and remain pinned to the matching heavy assets.\par
\textbf{Idempotent seeding.} Audit every \texttt{seed\_*()} function in \texttt{app.py} and \texttt{seed\_data.py}. Each function must return before any insert, update, or commit when its data already exists. Per-row guards are insufficient because a no-op commit can still change SQLite metadata. Boot from the proposed seed, run all seed functions again, reset, restart, and compare the complete runtime and seed database bytes after each operation.\par
\textbf{Benchmark state.} Seed the four standard users \texttt{alice.j@test.com}, \texttt{bob.c@test.com}, \texttt{carol.d@test.com}, and \texttt{david.k@test.com} with password \texttt{TestPass123!}. Give each account realistic profile data and the site-appropriate histories needed for authenticated tasks, including several saved items, cart or bag entries, prior orders or bookings, addresses, payment methods, subscriptions, reviews, or messages.\par
\textbf{Catalog and search.} Materialize runtime entities in SQLAlchemy rather than reading request-time JSON from \texttt{scraped\_data}. Populate approximately 50--200 primary entities, 5--15 categories, and multiple reviews for popular items where the domain supports them. Implement scored token-overlap search with stop-word removal and descending match score; do not require every query token to match. Pin a reference date when tasks depend on relative dates.\par
\textbf{Assets and provenance.} Keep \texttt{instance\_seed/*.db} and heavy image directories in the \ours HuggingFace dataset according to \texttt{.assetpaths}. Use \texttt{./scripts/extract\_assets.sh} to stage the asset change, review the dataset diff, upload it to the contributor's dataset fork, and obtain the merged asset revision. Update \texttt{.assets-revision} only after the HuggingFace change is merged. Cross-reference the asset and code changes; do not open either PR from this run unless explicitly requested.\par
\textbf{Final verification.} Run the asset checker and registry checker, build the image from the current checkout and pinned assets, start an isolated container, and require authenticated health for every registered site. Verify that the affected site serves all core routes, all key images decode, reset completes within the operational bound, every runtime database equals its seed after reset and startup, and a UI state change survives restart before the final reset.\par
\textbf{Deliverables.} Return the frozen seed database, asset diff and revision, database row counts by major model, benchmark-account state, scored-search result, build and health results, reset and restart byte checks, and exact remaining submission steps. Stop only after the local code and pinned assets reproduce the accepted environment deterministically.
\end{skillprompt}
\caption{Skill prompt used in Stage 5 to produce a deterministic and resettable seed database.}
\label{tab:prompt-seed-database}
\end{table}

%% file: tables/task_case_study.tex
\begin{table}[!t]
\centering
\footnotesize
\begin{tabular}{@{}>{\raggedright\arraybackslash}p{0.12\linewidth}p{0.04\linewidth}>{\raggedright\arraybackslash}p{0.18\linewidth}>{\raggedright\arraybackslash}p{0.56\linewidth}@{}}
\toprule
Website & ID & Workflow & Authenticated task \\
\midrule
\multirow{2}{*}{\textbf{Airbnb}} & 6 & Compare and book & Use dana.k@test.com / TestPass123! to help me choose from my 'Winter cabins' wishlist. Inspect both saved cabins, comparing their nightly rates, ratings and hot-tub amenities; check the second cabin's top review theme and count as well. Book the first saved cabin for two guests using its displayed dates, and report the selected name, booked dates, total and confirmation code alongside the comparison. \\
\cmidrule(lr){2-4}
 & 14 & Compare and book & Help me book a New York cultural tour for three guests using bob.c@test.com / TestPass123!. Compare the two cheapest Cultural tours by name, price per guest, minimum age and activity level. Choose the cheaper tour and book its first available date, then report the comparison, booked date, total and confirmation code. \\
\midrule
\multirow{2}{*}{\textbf{Carvana}} & 7 & Appraise and accept & Help me understand the demo valuation for VIN 1HGCM82633A004352 before accepting an offer. Compare Good and Excellent condition at 45,000 miles, and Good condition at 60,000 miles, reporting each simulated offer and the differences. Log in as bob.c@test.com with password TestPass123! and accept the 45,000-mile Good-condition offer. Confirm its saved offer code and amount. \\
\cmidrule(lr){2-4}
 & 15 & Reconcile and shortlist & I want to compare another Civic with the one I already ordered. Log in as alice.j@test.com with password TestPass123! and report my scheduled order's vehicle, monthly payment, term, APR and down payment. Find the first 2023 Civic in the Honda Civic results and estimate its payment with \$1,500 down over 60 months using Great credit. Save that alternative without cancelling my order, and report its price, amount financed, payment and the final saved-car list. \\
\midrule
\multirow{2}{*}{\textbf{Compass}} & 7 & Curate and share & Log in as david.kim@test.com (password webharbor123). Create a new collection called 'Austin top picks' and add exactly these two homes: 3305 Dolphin Drive and 1036 Liberty Park Drive, Unit 38A. Open the collection's share page to confirm both homes appear, then report the share token. \\
\cmidrule(lr){2-4}
 & 20 & Submit seller inquiry & Log in as alice.j@test.com (password webharbor123). Read the name, email, and phone currently shown on your account overview. Use those contact details to submit exactly one Sell With Us inquiry for ZIP code 94107 in this local Compass environment. Read the confirmation, then report its reference number and the submitted name, email, phone, and ZIP. Keep your account profile unchanged. \\
\bottomrule
\end{tabular}
\caption{Authenticated and multi-step tasks from five \ours environments (1/2, continued in Table~\ref{tab:task-case-study-2}). Each task requires authentication, a persistent state change, and cross-page reasoning.}
\label{tab:task-case-study}
\end{table}

\begin{table}[!t]
\centering
\footnotesize
\begin{tabular}{@{}>{\raggedright\arraybackslash}p{0.12\linewidth}p{0.04\linewidth}>{\raggedright\arraybackslash}p{0.18\linewidth}>{\raggedright\arraybackslash}p{0.56\linewidth}@{}}
\toprule
Website & ID & Workflow & Authenticated task \\
\midrule
\multirow{2}{*}{\textbf{Thumbtack}} & 12 & Review and message & Log in with the demo account (email: alice.j@test.com, password: TestPass123!). My house cleaning project is finished but I never left a review. Open the project, note the hired pro's price and their response note, and check their profile's current reviews. Then leave them a 5-star review saying they were thorough, including the word "spotless". Reopen their profile to confirm your review is the most recent one, and message them asking whether they could return for a move-out clean next month — follow up asking whether they'd bring their own supplies. Report both replies. \\
\cmidrule(lr){2-4}
 & 14 & Quote, hire, and review & Log in with the demo account (email: bob.c@test.com, password: TestPass123!). Ants have invaded my kitchen. Check the cost guide for what an exterminator typically runs, then find the exterminators who are Top Pros and request quotes for indoor ant treatment in zip 98101 within a week, describing the problem. Hire the responder with the most reviews, mark the project complete, and leave them a 5-star review mentioning the ants. Confirm on their profile that your review shows, and tell me their name, their quote, and the guide's typical range. \\
\midrule
\multirow{2}{*}{\textbf{YouTube}} & 2 & Subscribe and save & Sign in as carol.d@test.com using password TestPass123!. Open the AI Uncovered channel and inspect its public playlists and uploads. Open its most-viewed upload, subscribe to the channel and save that upload to Watch Later. Confirm the saved video and the new channel subscription on your account page. \\
\cmidrule(lr){2-4}
 & 18 & Comment and verify & Sign in as alice.j@test.com using password TestPass123!. Find Top 17 New Technology Trends That Will Define 2026 and open its watch page. Add the comment "Insightful breakdown of trend priorities.", save the video to Watch Later, then verify the exact new comment and the saved video in Watch Later. \\
\bottomrule
\end{tabular}
\caption{Authenticated and multi-step tasks from \ours environments (2/2, continued from Table~\ref{tab:task-case-study}).}
\label{tab:task-case-study-2}
\end{table}

%% file: tables/cross_benchmark_task_examples.tex
\begin{table}[!t]
\centering
\footnotesize
\begin{tabular}{@{}>{\raggedright\arraybackslash}p{0.19\linewidth}>{\raggedright\arraybackslash}p{0.77\linewidth}@{}}
\toprule
Benchmark / method & Representative released tasks \\
\midrule
{\raggedright WebArena~\citep{zhou2023webarena}\par} & {\raggedright \textbf{Task 1.} Go to the merge request on verification functions I have to review, find if the author of the merge request responded at the end, and reply "Thank you" if he did. Otherwise remind him with a simple @.\par \textbf{Task 2.} Create a folder named real\_space in gimmiethat.space repo. Within it, create a file named urls.txt that contains the URLs of the 5 most recent posts from the space?\par} \\
\midrule
{\raggedright Online-Mind2Web~\citep{xue2025illusion}\par} & {\raggedright \textbf{Task 1.} Find the weight of baggage allowance for economy class on Qatar Airways.\par \textbf{Task 2.} Find the cheapest used Honda Civic that meets all the following criteria: under \$25,000, fewer than 30,000 miles, includes Apple CarPlay, adaptive cruise control, and blind-spot sensors.\par} \\
\midrule
{\raggedright WebVoyager~\citep{he2024webvoyager}\par} & {\raggedright \textbf{Task 1.} Find a recipe for a vegan smoothie bowl on Allrecipes that includes bananas and leaves, has more than 20 reviews, and a rating of at least 4 stars. Provide a list of ingredients, preparation time, and a summary of the recipe steps.\par \textbf{Task 2.} Compare business class flight options from Lisbon to Singapore for a one-way trip on March 15, 2024, select one of the flights and see which websites offer its booking options. Which one is the cheapest.\par} \\
\midrule
{\raggedright CUA-Gym~\citep{wang2026cuagym}\par} & {\raggedright \textbf{Task 1.} Check all open PRs across our repos and report on CI status. For any PR with failing checks, add a comment tagging the author: '@\{author\} — CI is failing on this PR. Please check the \{check\_name\} step. Run details: \{detail\}'. Only comment on PRs that don't already have a CI-failure comment.\par \textbf{Task 2.} I'm preparing for a collaboration announcement. Create a new post with image https://picsum.photos/800/800?random=700, caption 'Big announcement coming soon! Stay tuned \#collaboration', location 'Downtown Studio'. Then send this post to the group chat with Alice and Bob, and also send a text message saying 'Excited to share this with you both!'\par} \\
\bottomrule
\end{tabular}
\caption{Representative released tasks from controlled, live, synthesized, and \ours suites (1/2, continued in Table~\ref{tab:cross-benchmark-task-examples-2}).}
\label{tab:cross-benchmark-task-examples}
\end{table}

\begin{table}[!t]
\centering
\footnotesize
\begin{tabular}{@{}>{\raggedright\arraybackslash}p{0.19\linewidth}>{\raggedright\arraybackslash}p{0.77\linewidth}@{}}
\toprule
Benchmark / method & Representative released tasks \\
\midrule
{\raggedright WebArena-Infinity~\citep{zhou2026webarenainfinity}\par} & {\raggedright \textbf{Task 1.} Approve the pharmacy's Atorvastatin-to-Rosuvastatin substitution, then prescribe Rosuvastatin 10mg tablet at bedtime for hyperlipidemia — qty 30, 5 refills, send to CVS \#4521.\par \textbf{Task 2.} Create a new invoice for Wellington \& Partners in NZD for 16 hours of consulting, and send it.\par} \\
\midrule
{\raggedright AutoWebWorld~\citep{wu2026autowebworld}\par} & {\raggedright \textbf{Task 1.} Book a flight from London Heathrow to New York JFK on March 17, 2028 in economy class, find American Airlines options, select the standard seat with standard baggage, fill out booking with first name Alex, last name Johnson, contact email alex.johnson@example.com, card number 4111111111111111, review and confirm the booking, then view the trip summary\par \textbf{Task 2.} Log into my account using email user@example.com and password SecurePass123, then go to payment methods, select PayPal, authorize payment with PayPal email paypaluser@example.com\par} \\
\midrule
{\raggedright \ours\par} & {\raggedright \textbf{Task 1.} Sign in as alice.j@test.com using password TestPass123!. Find Top 17 New Technology Trends That Will Define 2026 and open its watch page. Add the comment "Insightful breakdown of trend priorities.", save the video to Watch Later, then verify the exact new comment and the saved video in Watch Later.\par \textbf{Task 2.} Log in as david.kim@test.com (password webharbor123). Create a new collection called 'Austin top picks' and add exactly these two homes: 3305 Dolphin Drive and 1036 Liberty Park Drive, Unit 38A. Open the collection's share page to confirm both homes appear, then report the share token.\par} \\
\bottomrule
\end{tabular}
\caption{Representative released tasks (2/2, continued from Table~\ref{tab:cross-benchmark-task-examples}).}
\label{tab:cross-benchmark-task-examples-2}
\end{table}

%% file: figures/cmp_env_case_study_youtube.tex
\begin{figure}[!t]
\centering
\includegraphics[width=0.84\linewidth]{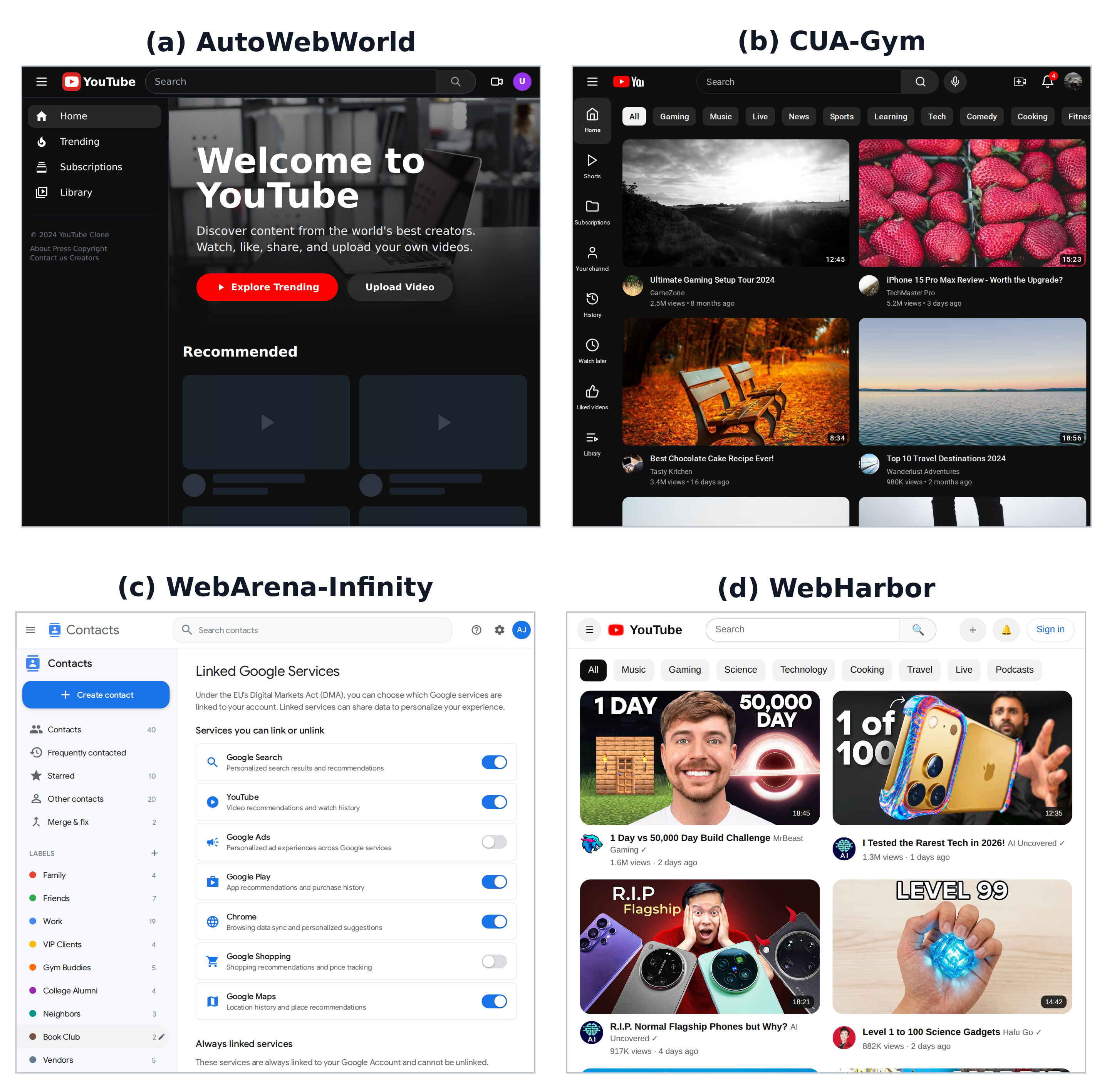}
\caption{Comparison of YouTube-related environments. (a) AutoWebWorld uses a welcome template with placeholder recommendations. (b) CUA-Gym reproduces the video-feed structure with generic thumbnails and irrelevant images. (c) WebArena-Infinity has no YouTube website. Its closest YouTube-specific page is an image-free Linked Services template in the Gmail Accounts \& Contacts environment. (d) \ours provides target-like navigation, video-specific thumbnails, channel identity, and engagement metadata such as view counts.}
\label{fig:cmp-env-case-study-youtube}
\end{figure}

%% file: figures/our_envs_gallery.tex
\begin{figure}[!t]
\centering
\captionsetup[subfigure]{font=footnotesize,labelformat=empty,skip=1pt}
\begin{subfigure}[b]{0.32\textwidth}
\includegraphics[width=\linewidth]{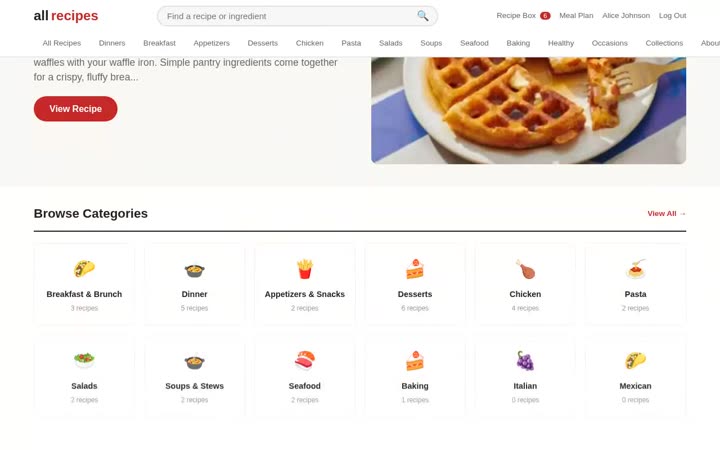}
\caption{Allrecipes}
\end{subfigure}\hfill
\begin{subfigure}[b]{0.32\textwidth}
\includegraphics[width=\linewidth]{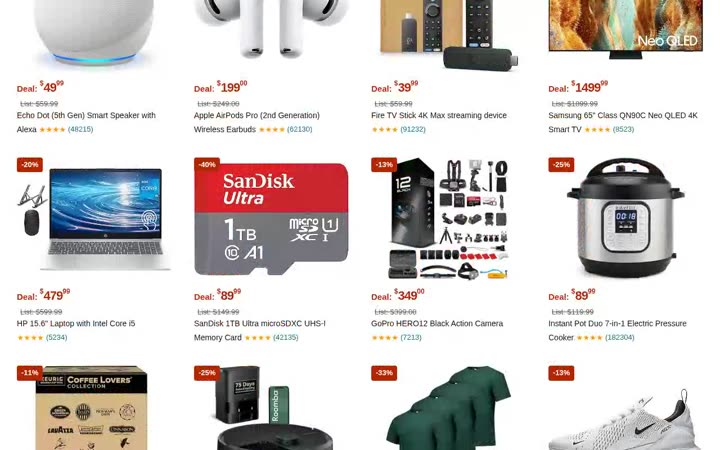}
\caption{Amazon}
\end{subfigure}\hfill
\begin{subfigure}[b]{0.32\textwidth}
\includegraphics[width=\linewidth]{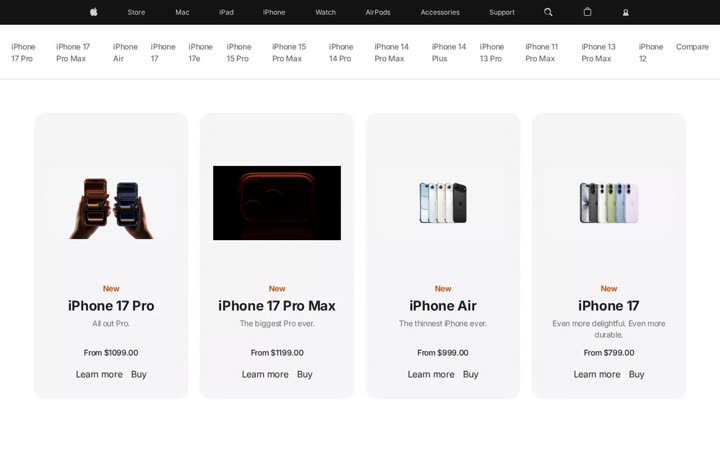}
\caption{Apple}
\end{subfigure}
\par\vspace{4pt}
\begin{subfigure}[b]{0.32\textwidth}
\includegraphics[width=\linewidth]{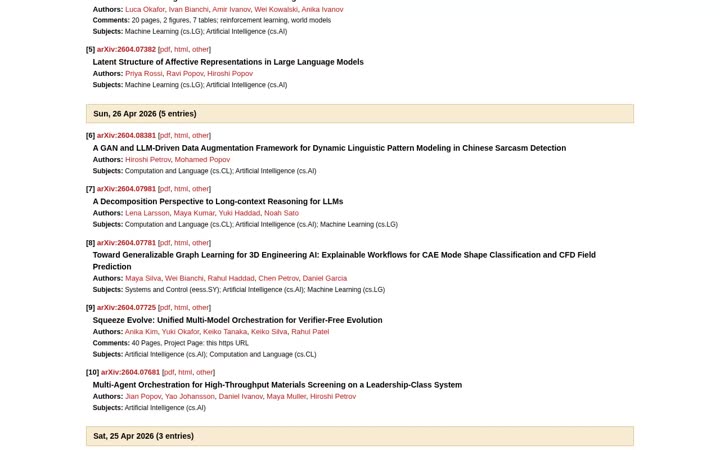}
\caption{ArXiv}
\end{subfigure}\hfill
\begin{subfigure}[b]{0.32\textwidth}
\includegraphics[width=\linewidth]{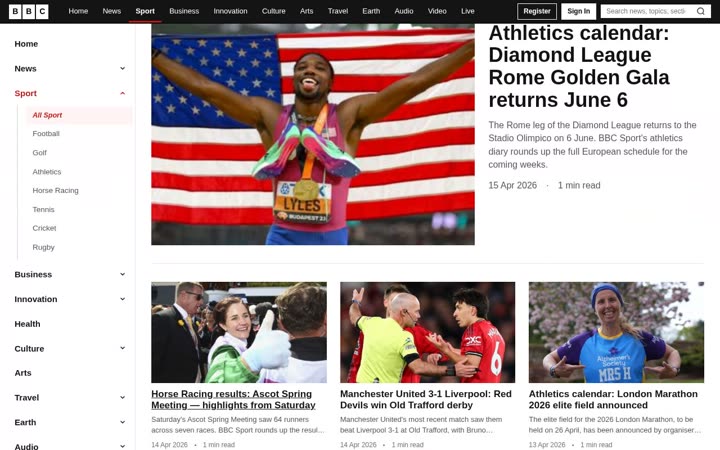}
\caption{BBC News}
\end{subfigure}\hfill
\begin{subfigure}[b]{0.32\textwidth}
\includegraphics[width=\linewidth]{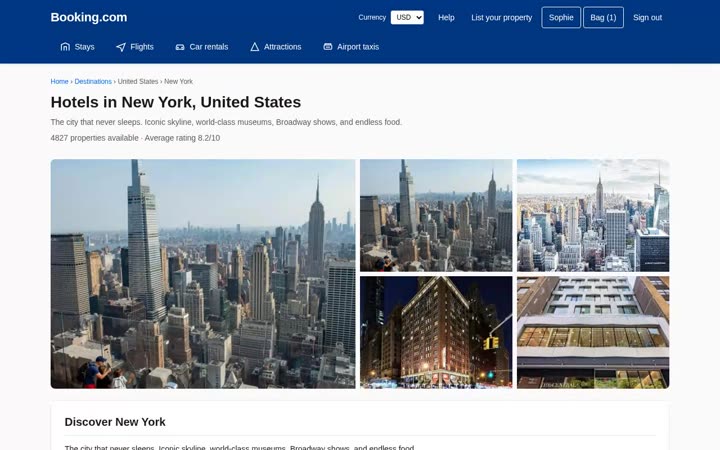}
\caption{Booking}
\end{subfigure}
\par\vspace{4pt}
\begin{subfigure}[b]{0.32\textwidth}
\includegraphics[width=\linewidth]{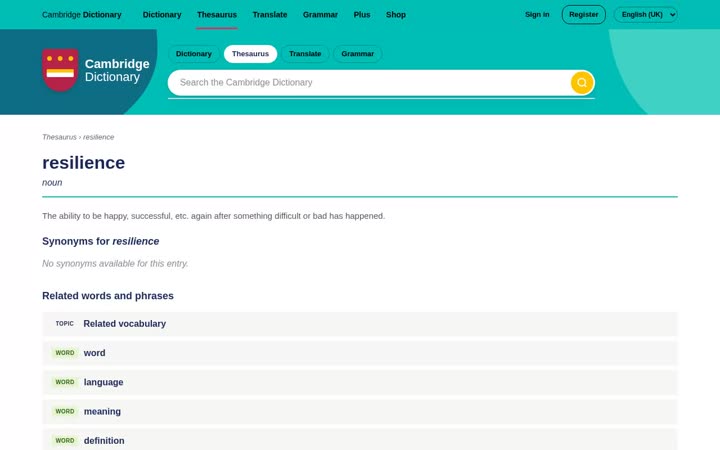}
\caption{Cambridge Dict.}
\end{subfigure}\hfill
\begin{subfigure}[b]{0.32\textwidth}
\includegraphics[width=\linewidth]{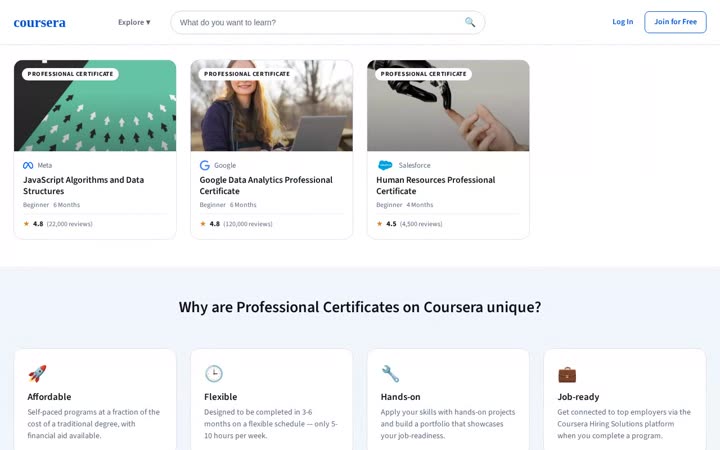}
\caption{Coursera}
\end{subfigure}\hfill
\begin{subfigure}[b]{0.32\textwidth}
\includegraphics[width=\linewidth]{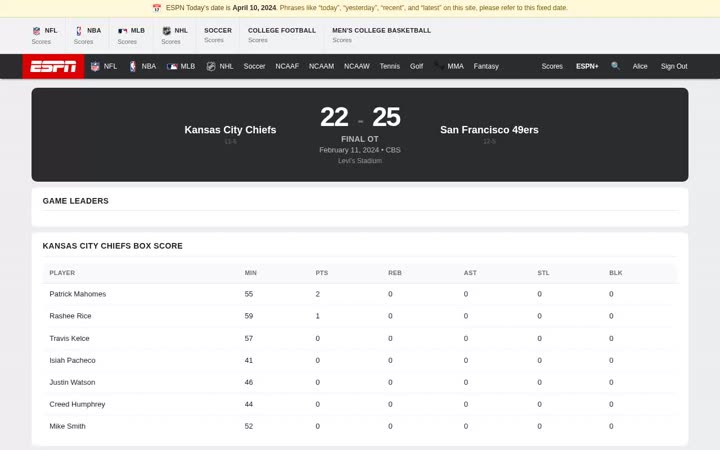}
\caption{ESPN}
\end{subfigure}
\par\vspace{4pt}
\begin{subfigure}[b]{0.32\textwidth}
\includegraphics[width=\linewidth]{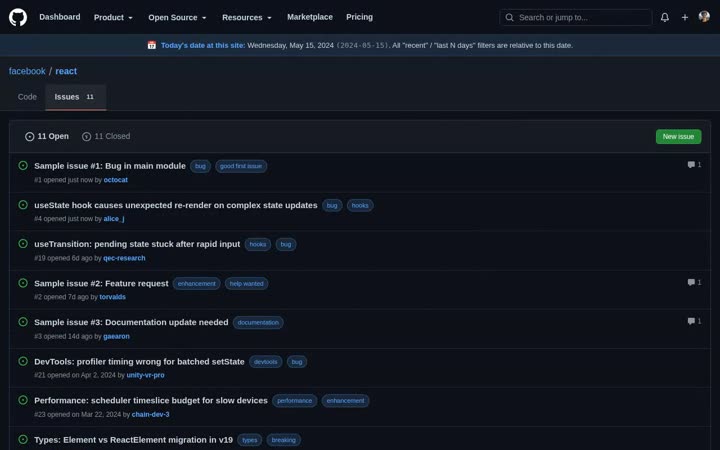}
\caption{GitHub}
\end{subfigure}\hfill
\begin{subfigure}[b]{0.32\textwidth}
\includegraphics[width=\linewidth]{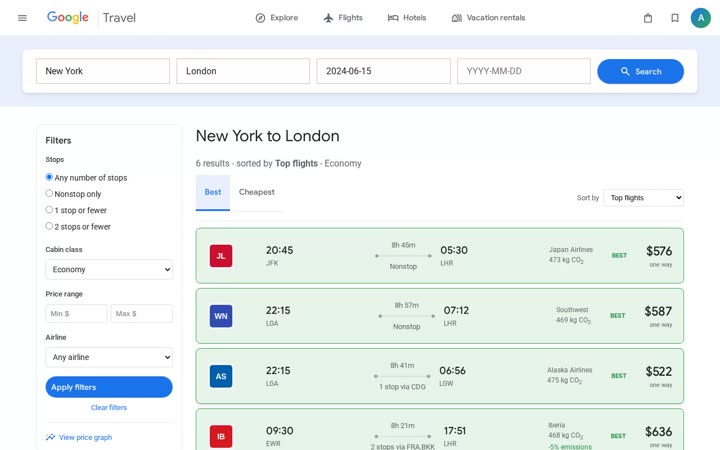}
\caption{Google Flights}
\end{subfigure}\hfill
\begin{subfigure}[b]{0.32\textwidth}
\includegraphics[width=\linewidth]{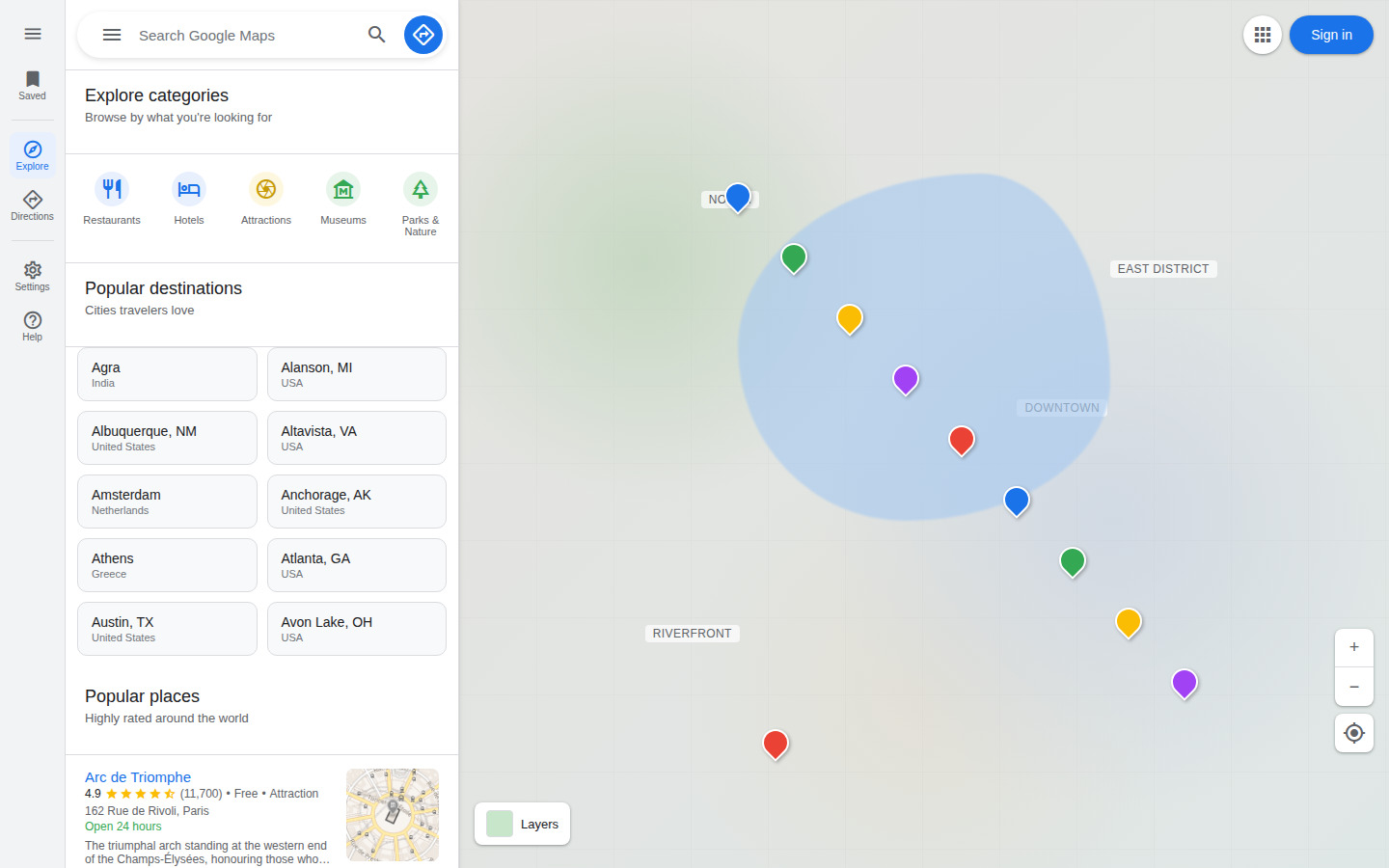}
\caption{Google Maps}
\end{subfigure}
\par\vspace{4pt}
\begin{subfigure}[b]{0.32\textwidth}
\includegraphics[width=\linewidth]{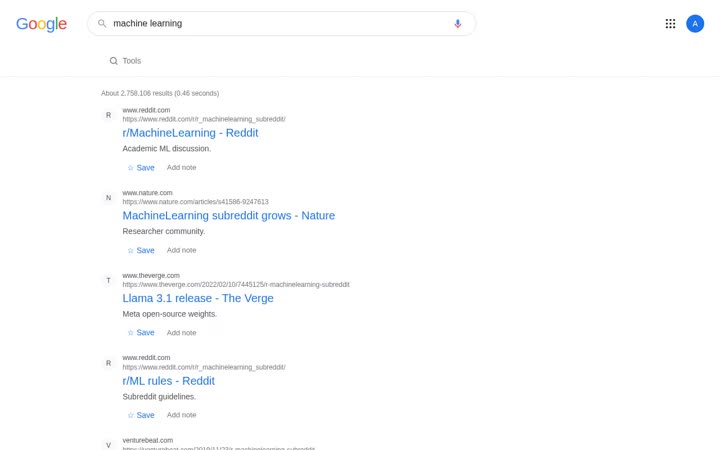}
\caption{Google Search}
\end{subfigure}\hfill
\begin{subfigure}[b]{0.32\textwidth}
\includegraphics[width=\linewidth]{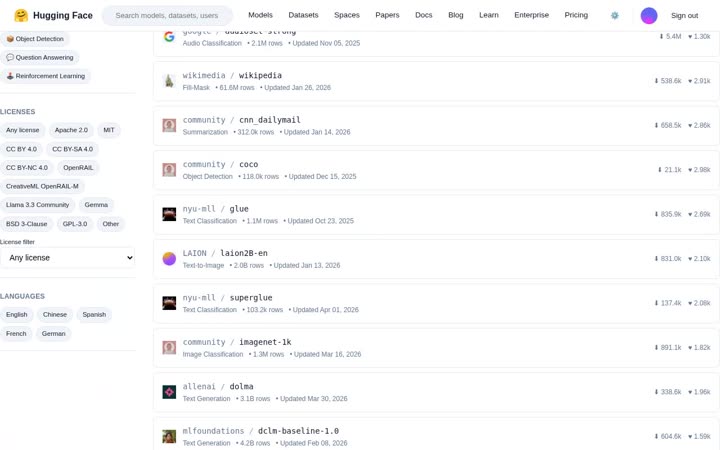}
\caption{Hugging Face}
\end{subfigure}\hfill
\begin{subfigure}[b]{0.32\textwidth}
\includegraphics[width=\linewidth]{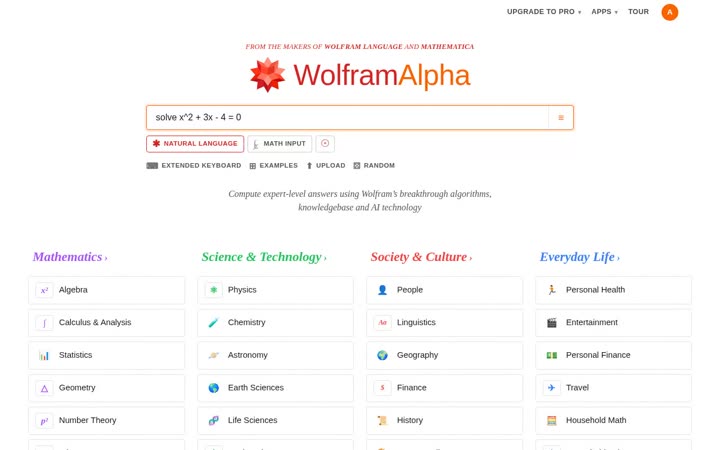}
\caption{Wolfram Alpha}
\end{subfigure}
\caption{Representative pages from 30 \ours environments across commerce, travel, media, entertainment, developer tools, and information services (1/2, continued in Figure~\ref{fig:gallery_full-2}). The examples show the page structures, imagery, navigation, metadata, and controls implemented across the suite.}
\label{fig:gallery_full}
\end{figure}

\begin{figure}[!t]
\centering
\captionsetup[subfigure]{font=footnotesize,labelformat=empty,skip=1pt}
\begin{subfigure}[b]{0.32\textwidth}
\includegraphics[width=\linewidth]{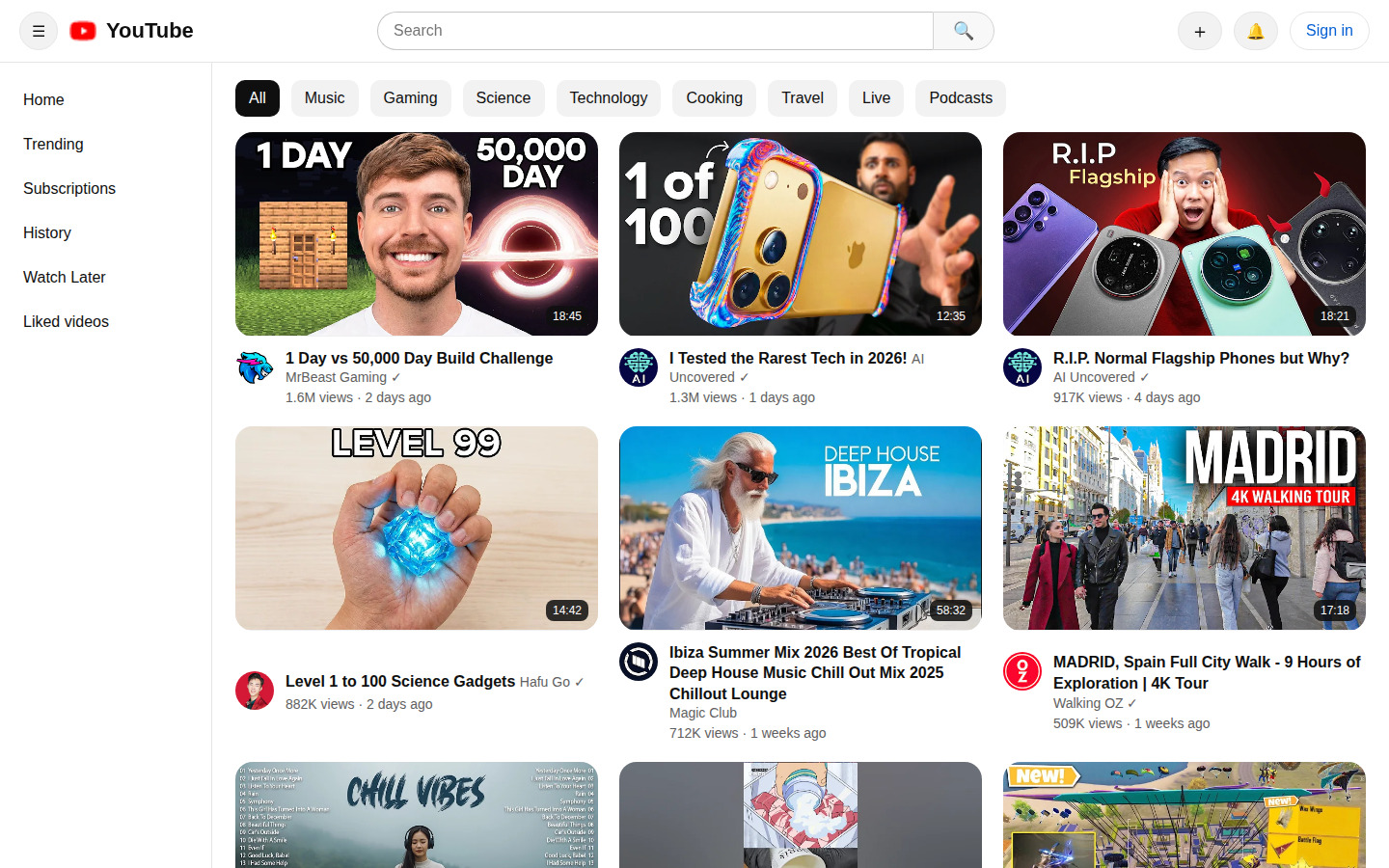}
\caption{YouTube}
\end{subfigure}\hfill
\begin{subfigure}[b]{0.32\textwidth}
\includegraphics[width=\linewidth]{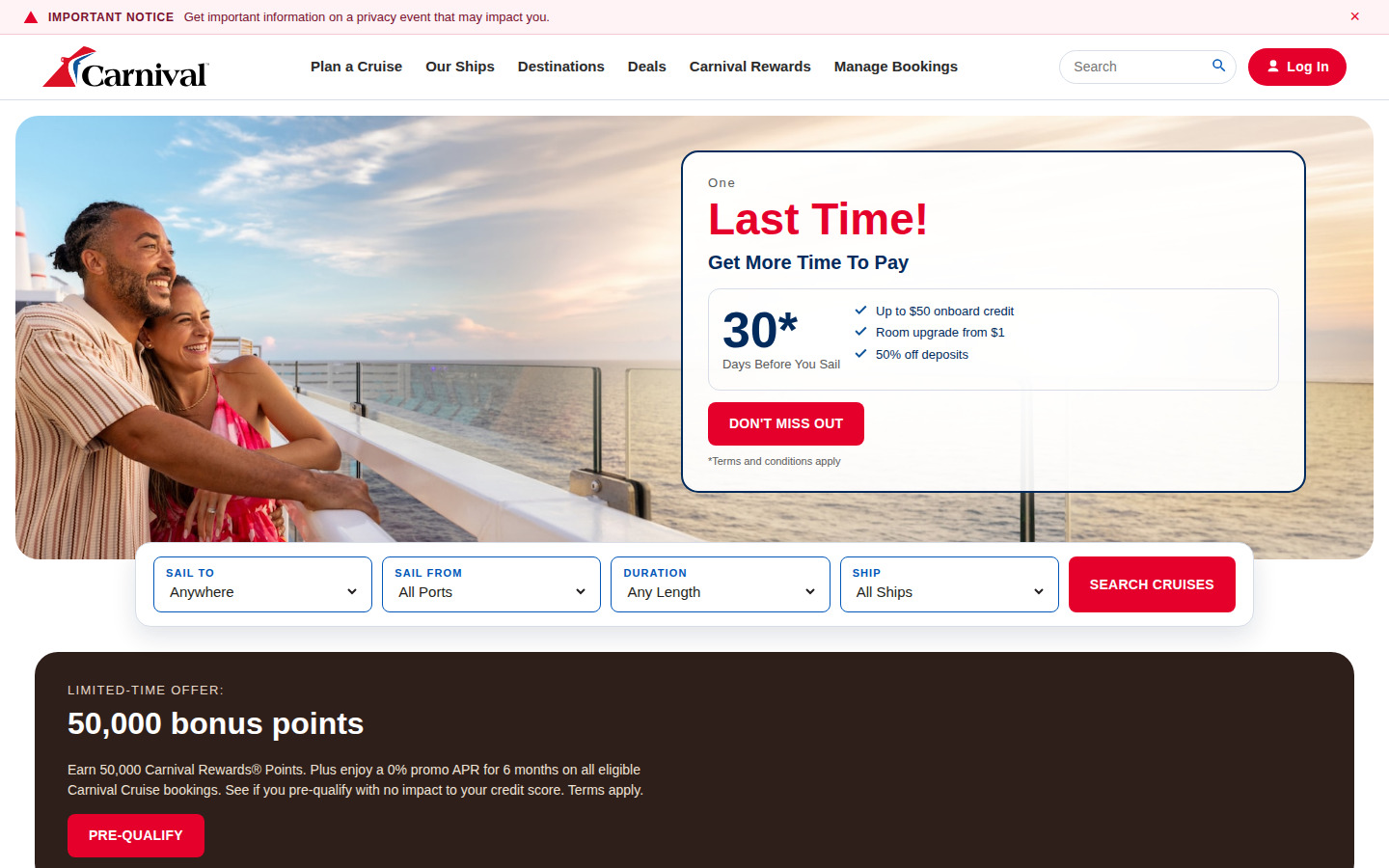}
\caption{Carnival Cruise}
\end{subfigure}\hfill
\begin{subfigure}[b]{0.32\textwidth}
\includegraphics[width=\linewidth]{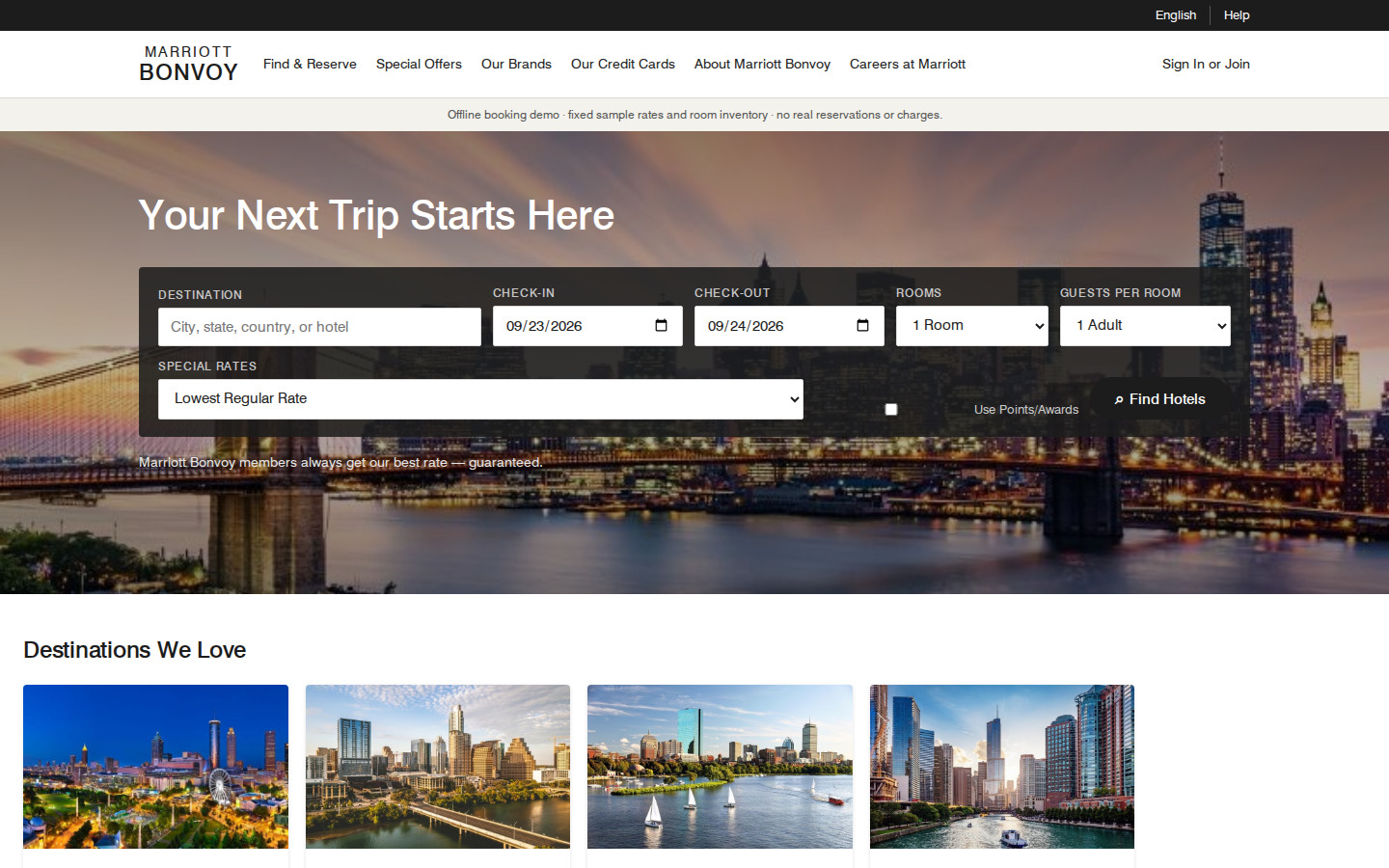}
\caption{Marriott}
\end{subfigure}
\par\vspace{4pt}
\begin{subfigure}[b]{0.32\textwidth}
\includegraphics[width=\linewidth]{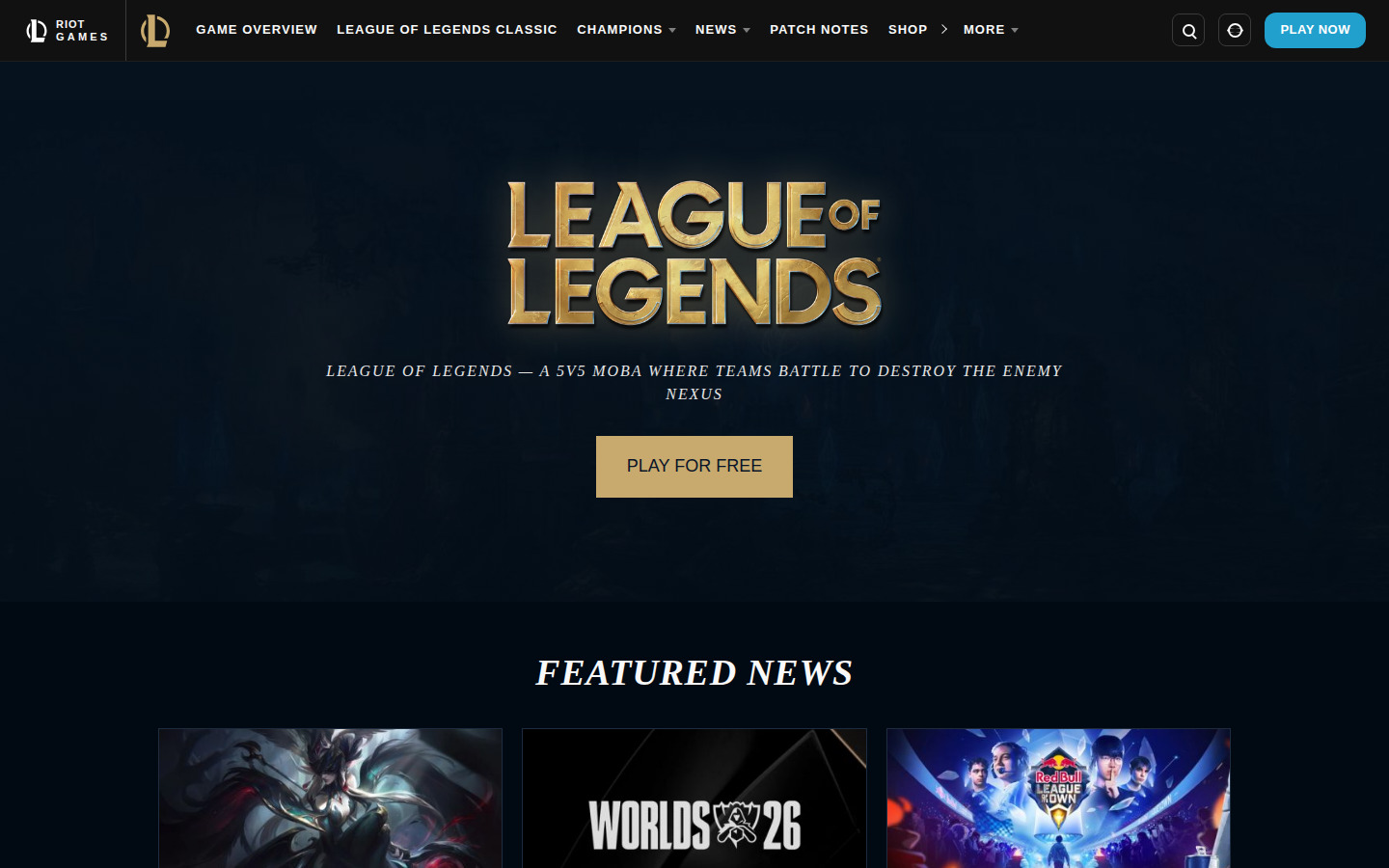}
\caption{League of Legends}
\end{subfigure}\hfill
\begin{subfigure}[b]{0.32\textwidth}
\includegraphics[width=\linewidth]{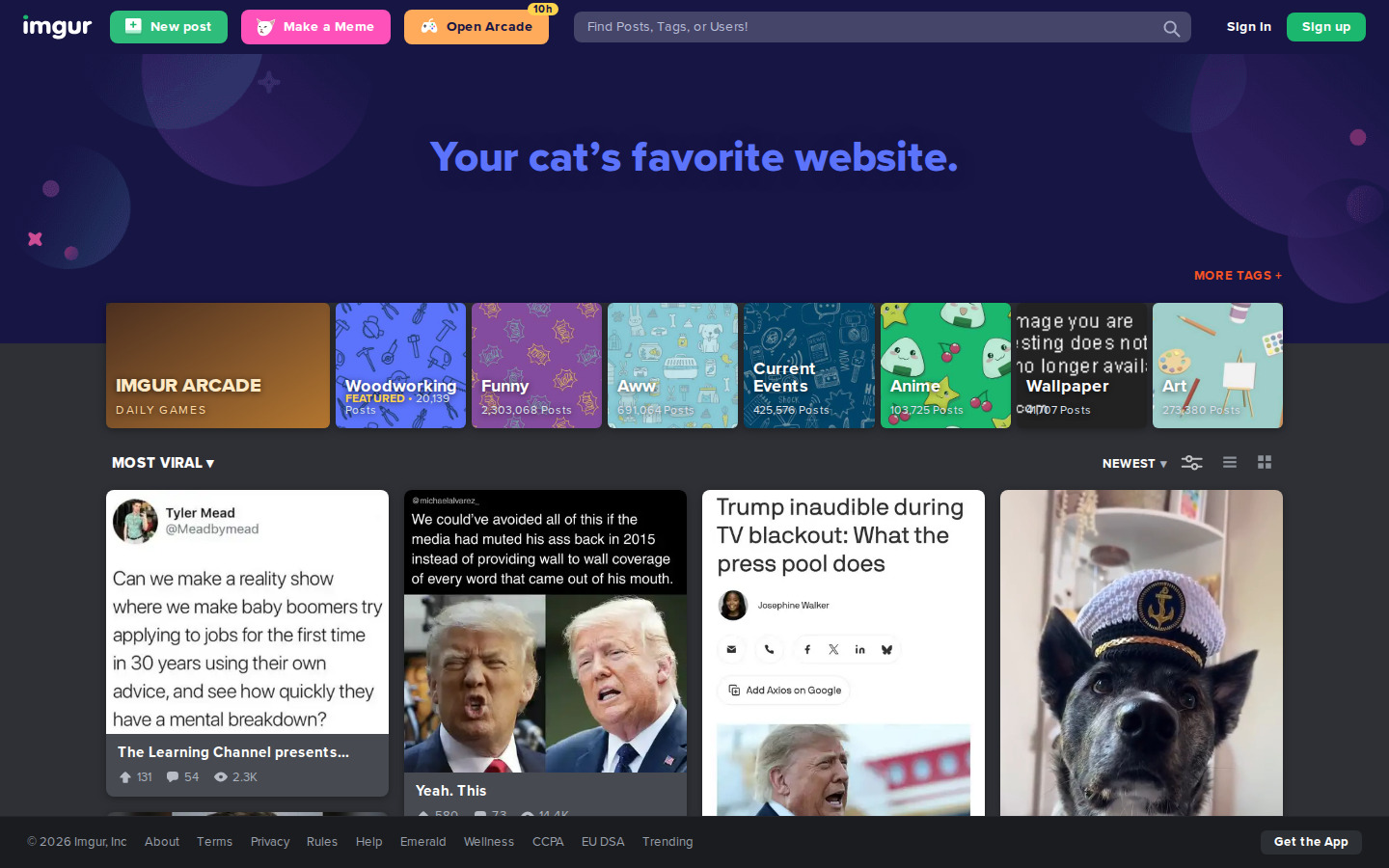}
\caption{Imgur}
\end{subfigure}\hfill
\begin{subfigure}[b]{0.32\textwidth}
\includegraphics[width=\linewidth]{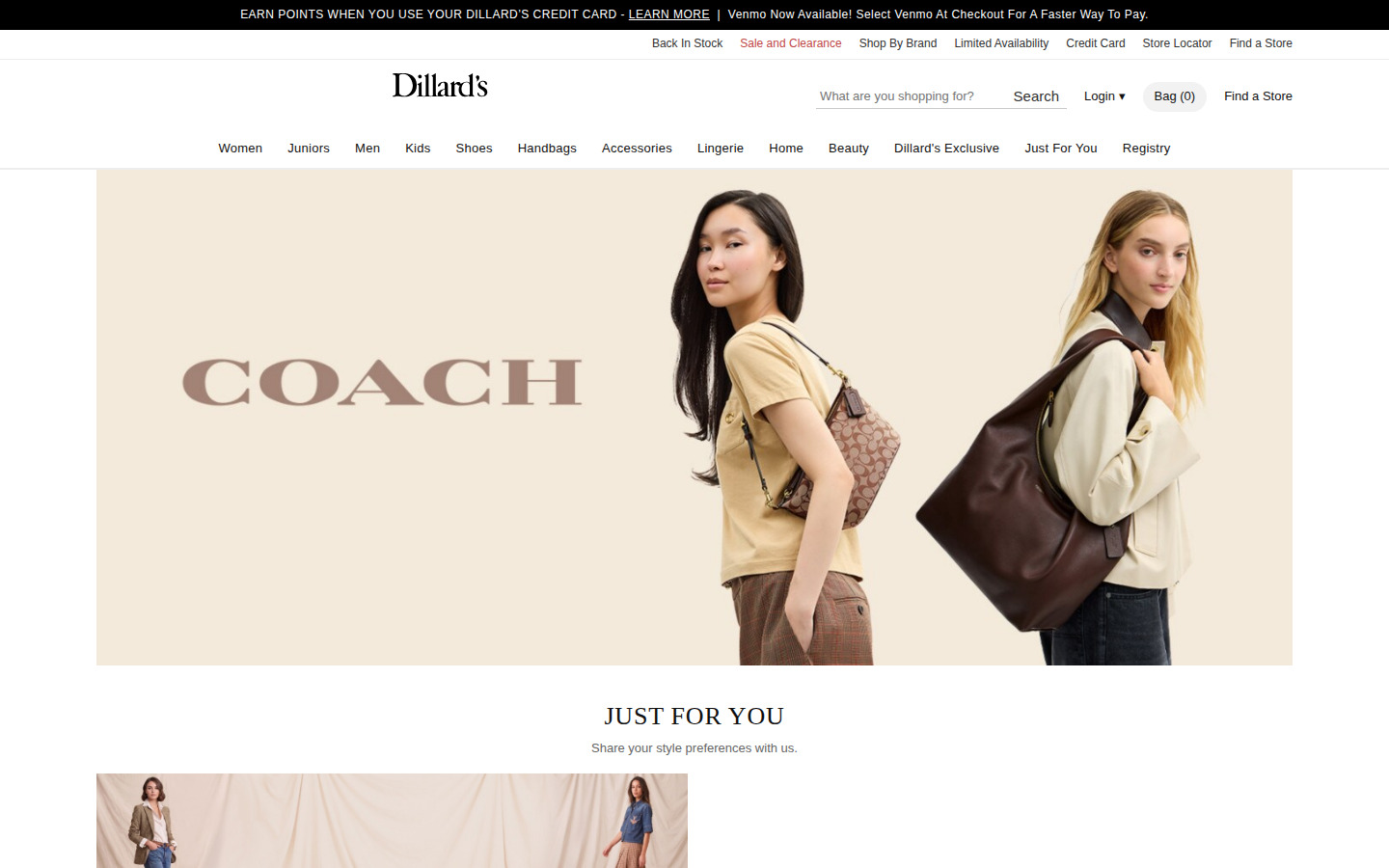}
\caption{Dillard's}
\end{subfigure}
\par\vspace{4pt}
\begin{subfigure}[b]{0.32\textwidth}
\includegraphics[width=\linewidth]{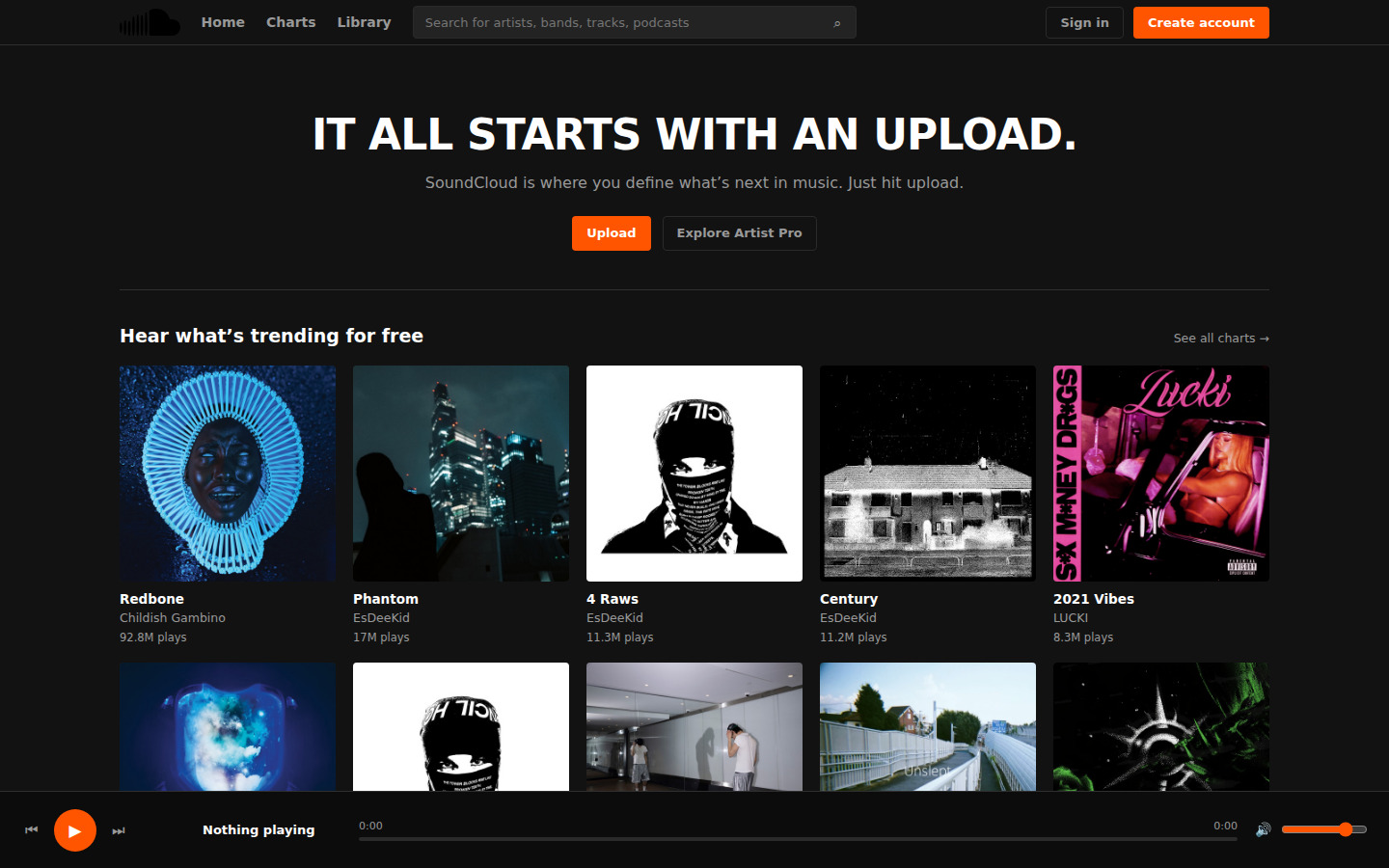}
\caption{SoundCloud}
\end{subfigure}\hfill
\begin{subfigure}[b]{0.32\textwidth}
\includegraphics[width=\linewidth]{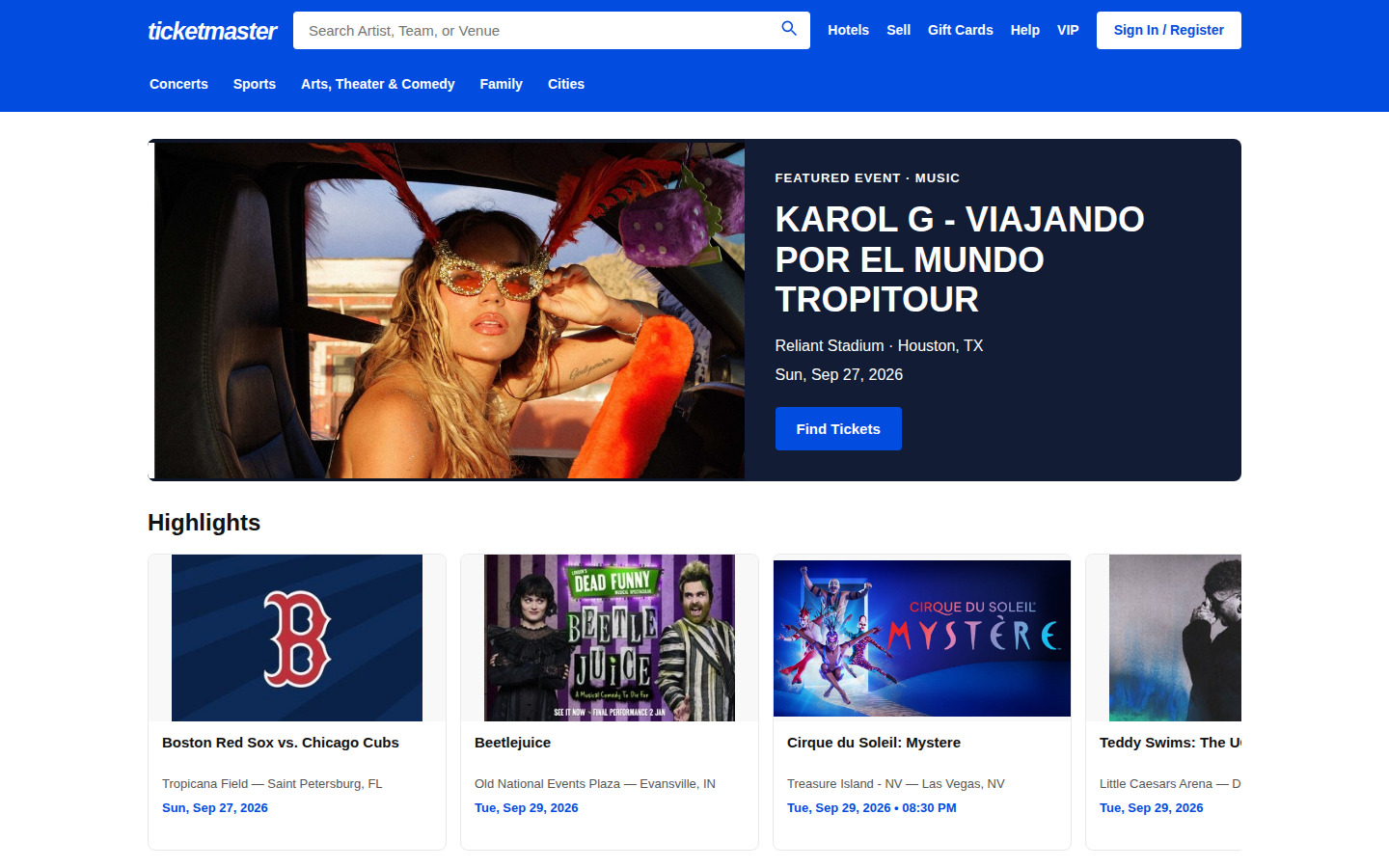}
\caption{Ticketmaster}
\end{subfigure}\hfill
\begin{subfigure}[b]{0.32\textwidth}
\includegraphics[width=\linewidth]{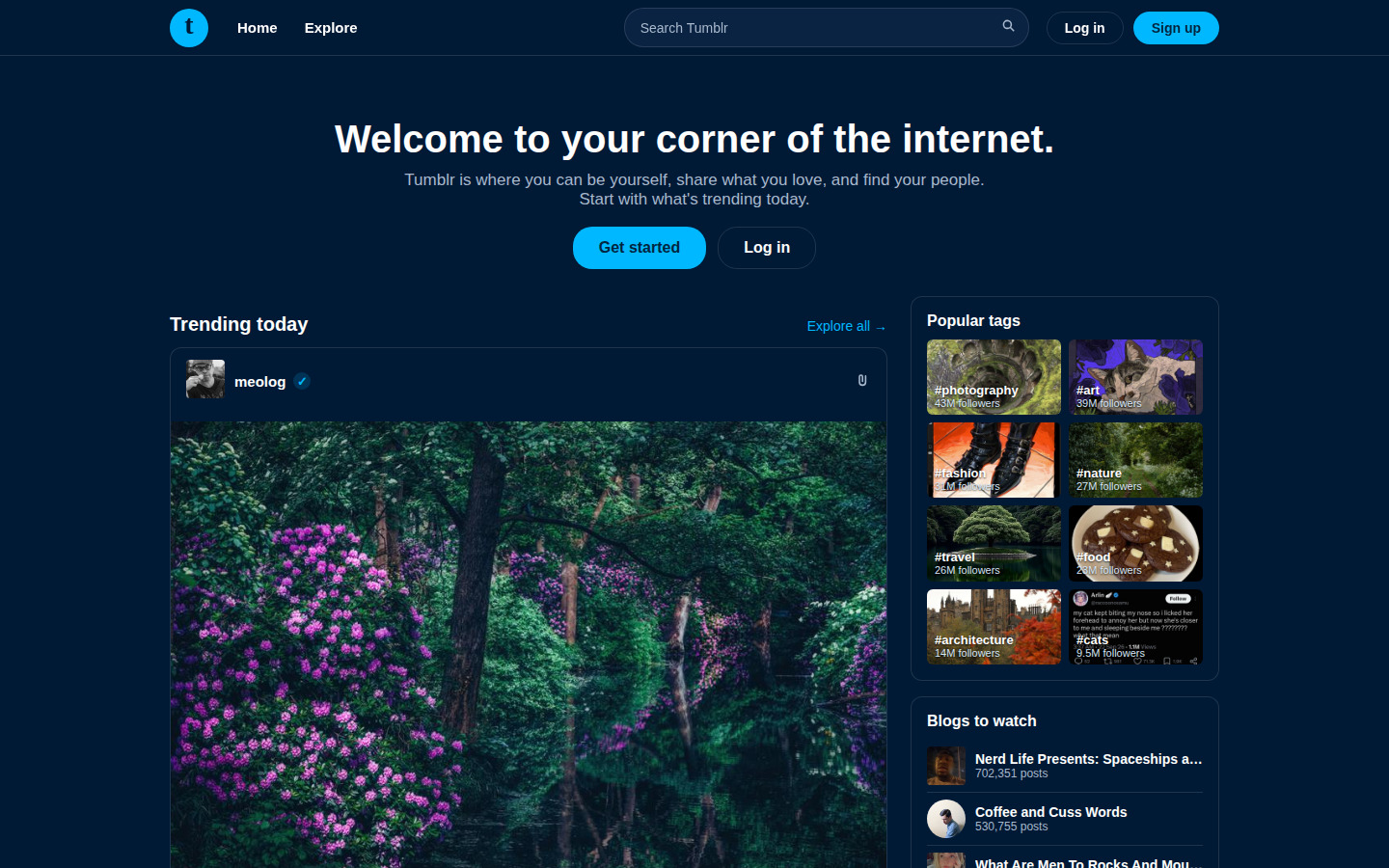}
\caption{Tumblr}
\end{subfigure}
\par\vspace{4pt}
\begin{subfigure}[b]{0.32\textwidth}
\includegraphics[width=\linewidth]{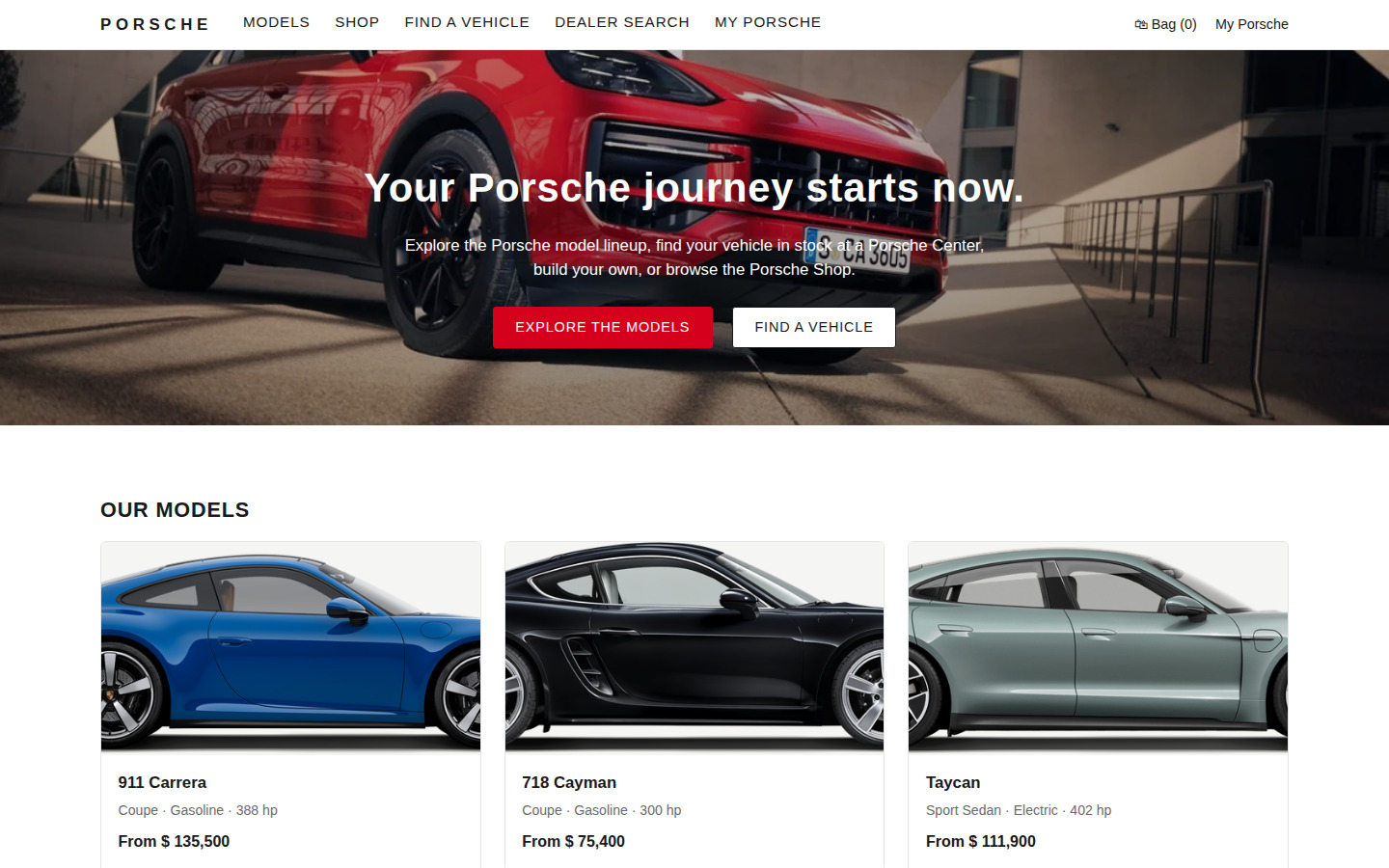}
\caption{Porsche}
\end{subfigure}\hfill
\begin{subfigure}[b]{0.32\textwidth}
\includegraphics[width=\linewidth]{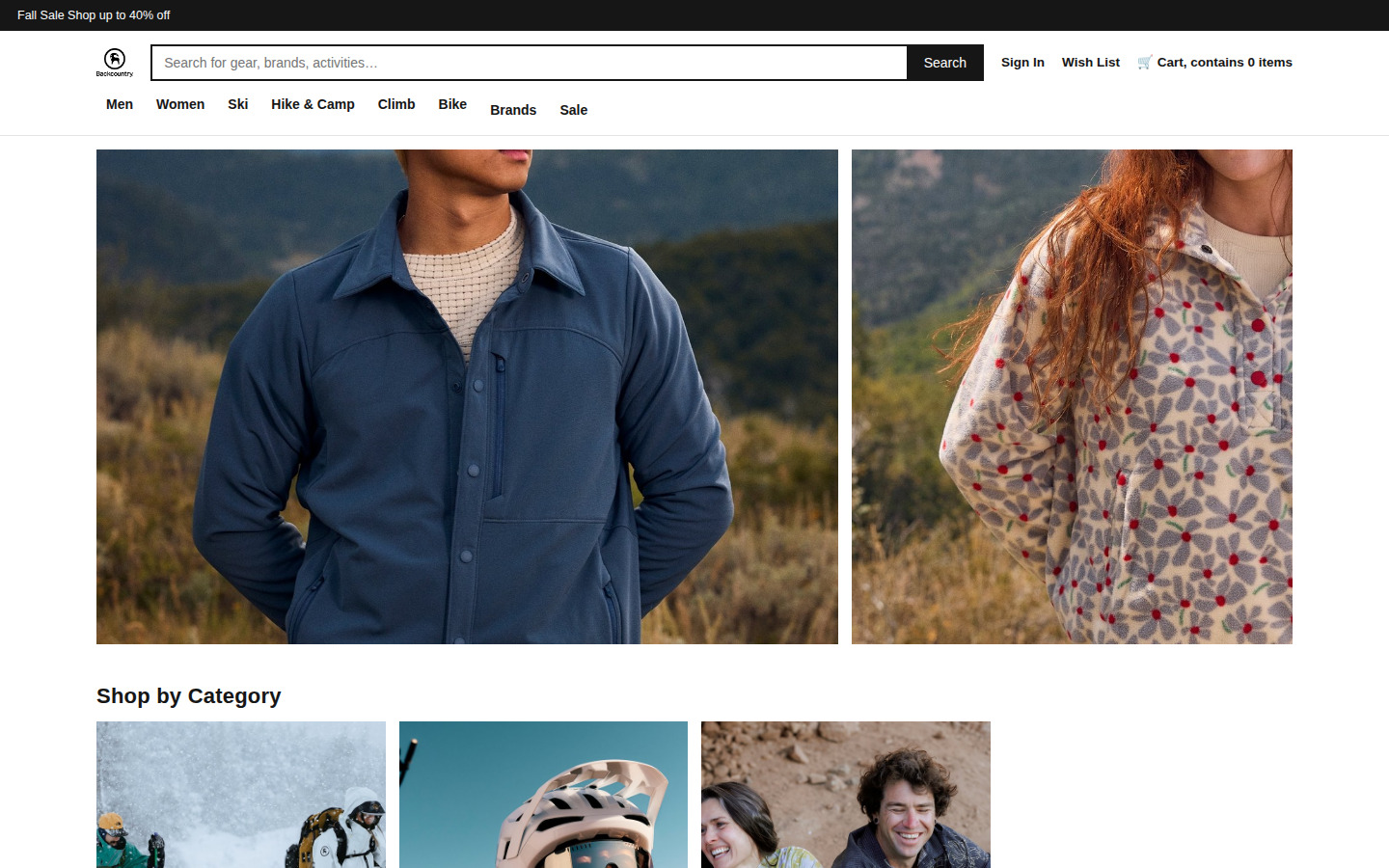}
\caption{Backcountry}
\end{subfigure}\hfill
\begin{subfigure}[b]{0.32\textwidth}
\includegraphics[width=\linewidth]{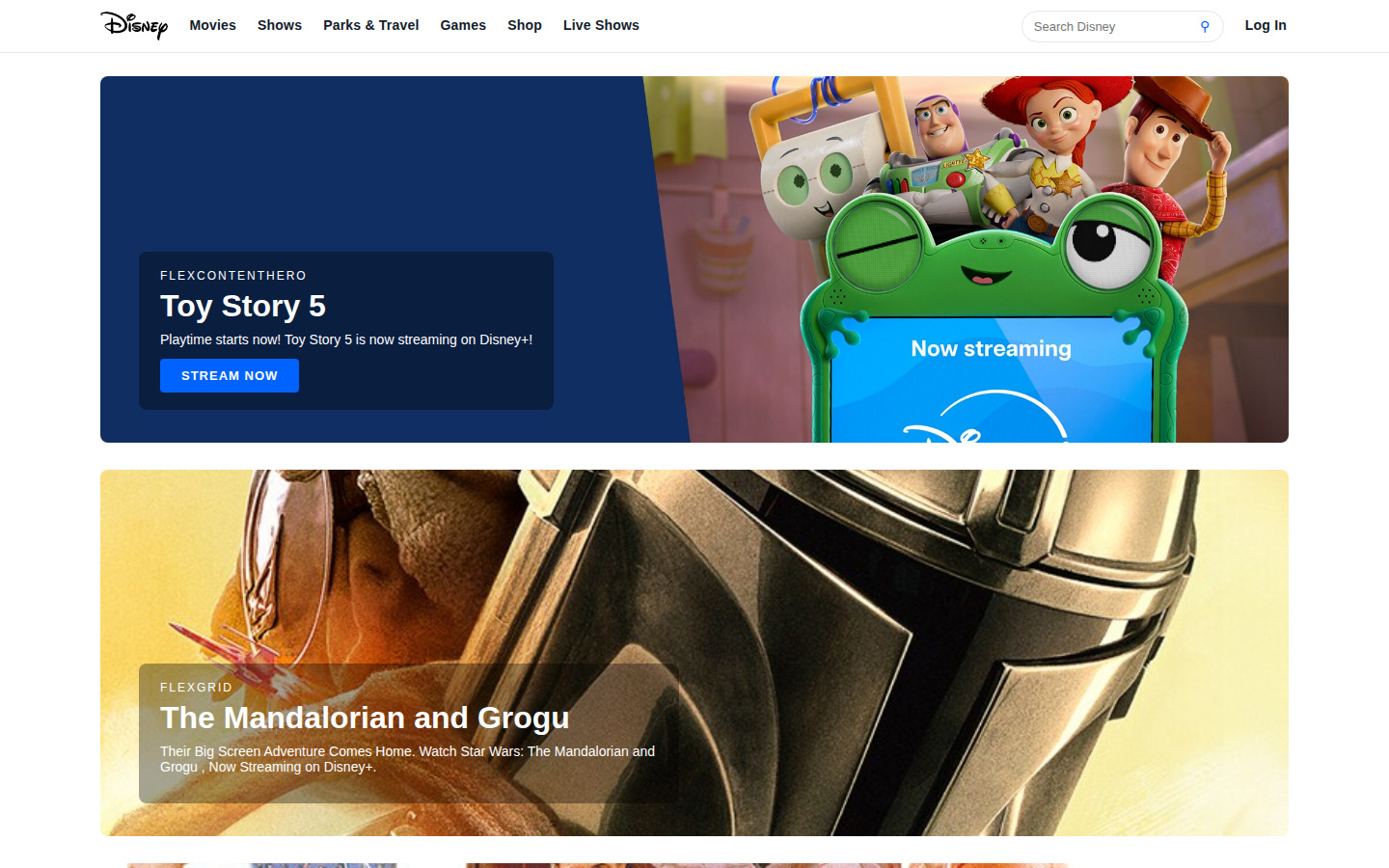}
\caption{Disney}
\end{subfigure}
\par\vspace{4pt}
\begin{subfigure}[b]{0.32\textwidth}
\includegraphics[width=\linewidth]{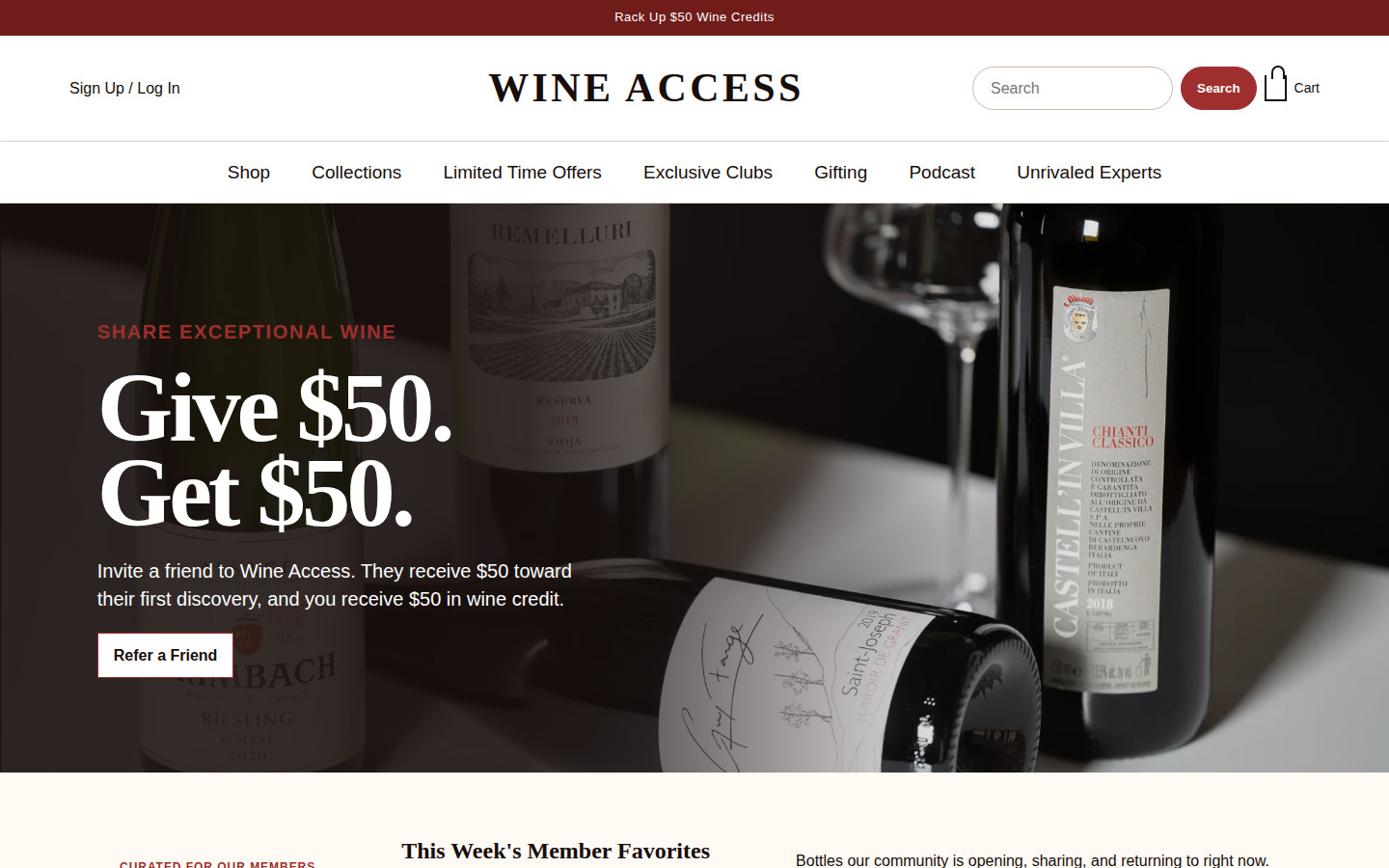}
\caption{Wine Access}
\end{subfigure}\hfill
\begin{subfigure}[b]{0.32\textwidth}
\includegraphics[width=\linewidth]{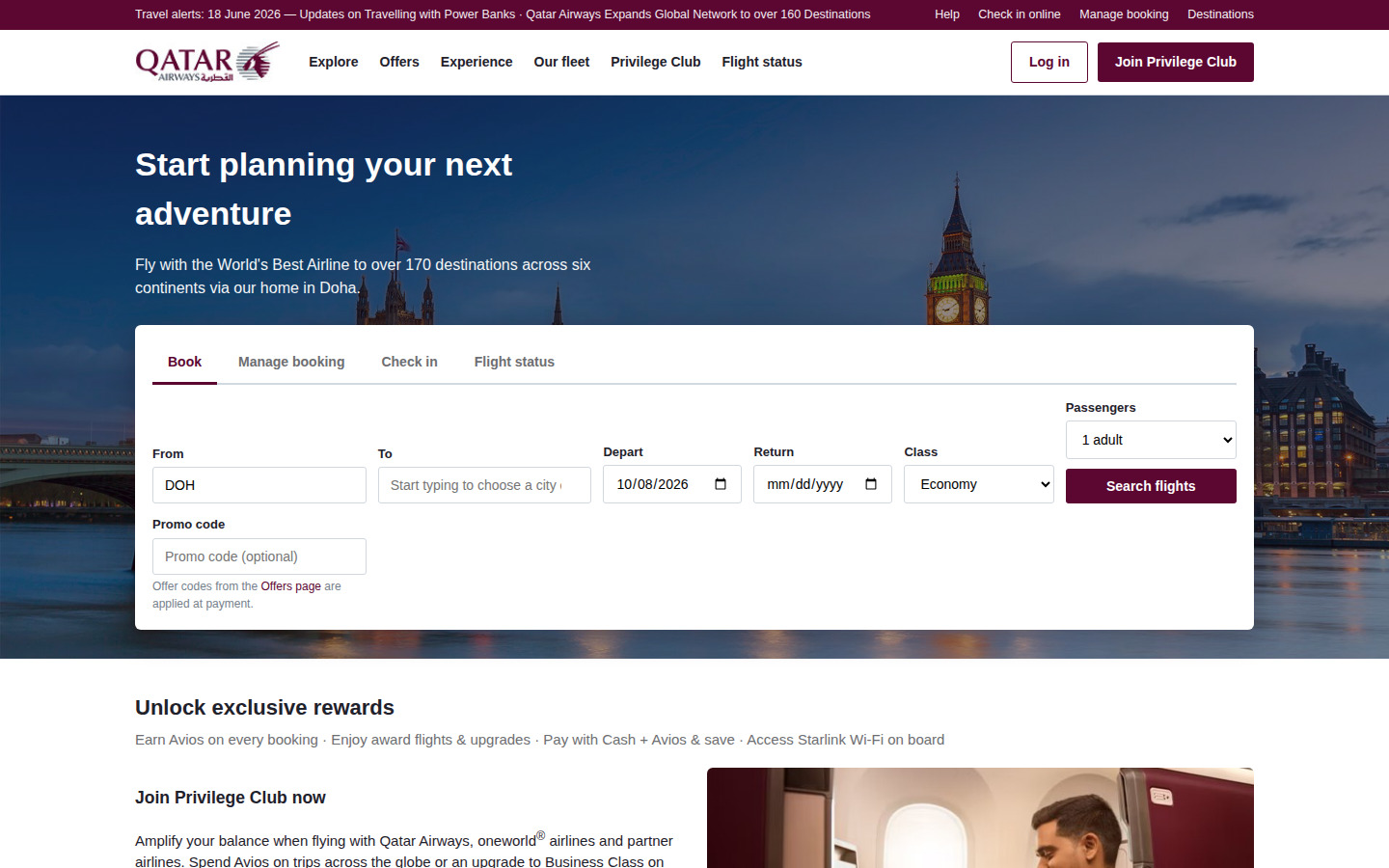}
\caption{Qatar Airways}
\end{subfigure}\hfill
\begin{subfigure}[b]{0.32\textwidth}
\includegraphics[width=\linewidth]{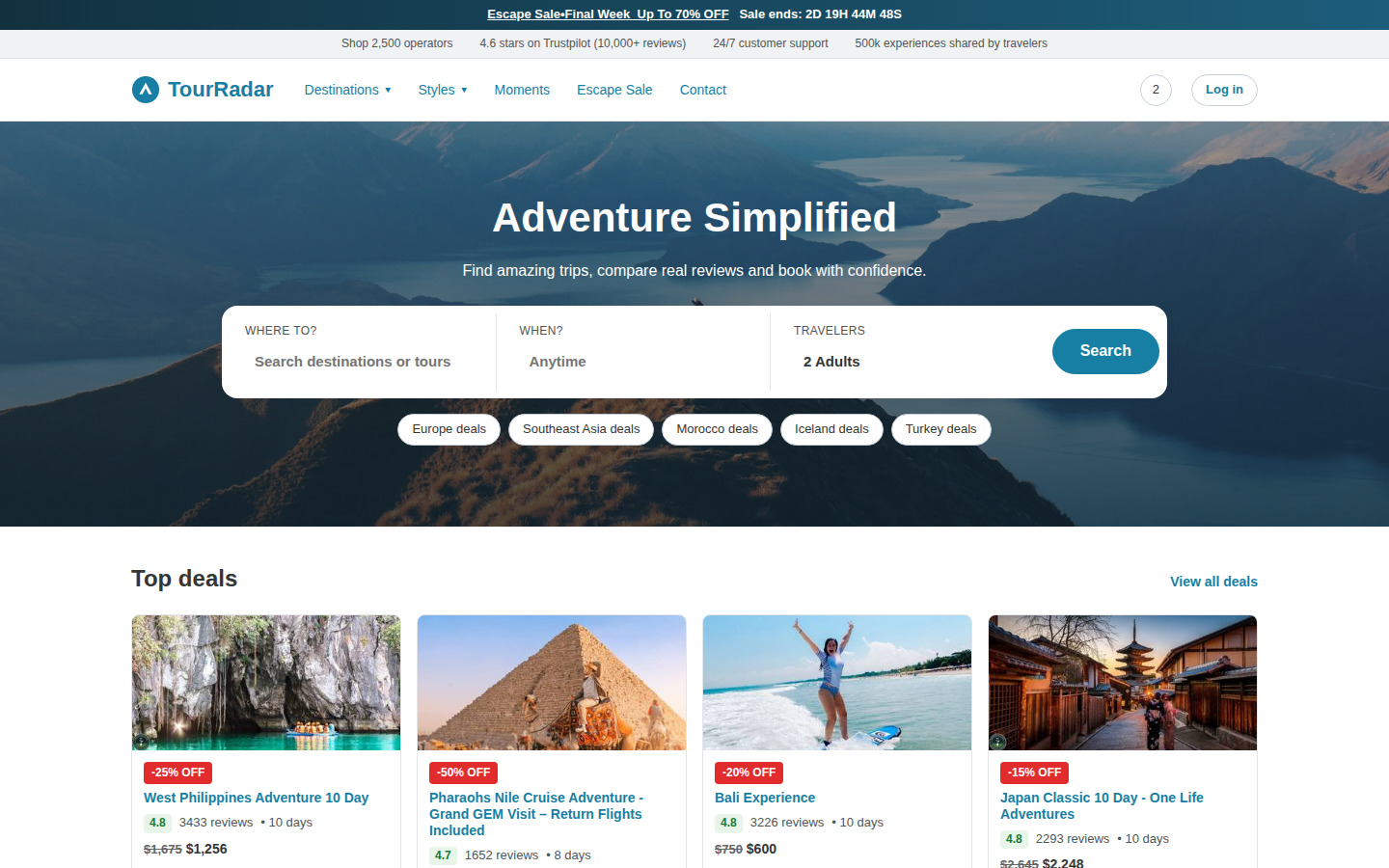}
\caption{TourRadar}
\end{subfigure}
\caption{Representative pages from \ours environments (2/2, continued from Figure~\ref{fig:gallery_full}).}
\label{fig:gallery_full-2}
\end{figure}

%% file: tables/full_envs_list.tex
\begin{table}[p]
\centering
\footnotesize
\setlength{\tabcolsep}{3pt}
\begin{tabular}{@{}>{\raggedright\arraybackslash}p{0.35\linewidth}cc@{\quad}>{\raggedright\arraybackslash}p{0.35\linewidth}cc@{}}
\toprule
Environment & WV & OM2W & Environment & WV & OM2W \\
\midrule
4shared & -- & $\bullet$ & Coolmath4Kids & -- & $\bullet$ \\
9GAG & -- & $\bullet$ & Coursera & $\bullet$ & $\bullet$ \\
AccuWeather & -- & $\bullet$ & Craigslist & -- & $\bullet$ \\
Adopt a Pet & -- & $\bullet$ & CVS & -- & $\bullet$ \\
Airbnb & -- & $\bullet$ & dblp & -- & $\bullet$ \\
AKC & -- & $\bullet$ & Dillards & -- & $\bullet$ \\
Allrecipes & $\bullet$ & $\bullet$ & Discogs & -- & $\bullet$ \\
Amazon & $\bullet$ & $\bullet$ & Disney & -- & $\bullet$ \\
Amazon Jobs & -- & $\bullet$ & Drugs.com & -- & $\bullet$ \\
America's Health Rankings & -- & $\bullet$ & ESPN & $\bullet$ & $\bullet$ \\
American Express & -- & $\bullet$ & Eventbrite & -- & $\bullet$ \\
Amtrak & -- & $\bullet$ & Fandom & -- & $\bullet$ \\
Apartments.com & -- & $\bullet$ & FedEx & -- & $\bullet$ \\
Apple & $\bullet$ & $\bullet$ & FlightAware & -- & $\bullet$ \\
ArXiv & $\bullet$ & $\bullet$ & FOX Sports & -- & $\bullet$ \\
B\&H Photo & -- & $\bullet$ & GitHub & $\bullet$ & $\bullet$ \\
BabyCenter & -- & $\bullet$ & Google Finance & -- & $\bullet$ \\
Backcountry & -- & $\bullet$ & Google Flights & $\bullet$ & -- \\
Bandcamp & -- & $\bullet$ & Google Map & $\bullet$ & $\bullet$ \\
BBC News & $\bullet$ & -- & Google Search & $\bullet$ & -- \\
Best Buy & -- & $\bullet$ & Google Shopping & -- & $\bullet$ \\
Better Business Bureau & -- & $\bullet$ & GOV.UK & -- & $\bullet$ \\
Birkenstock & -- & $\bullet$ & Healthgrades & -- & $\bullet$ \\
BoardGameGeek & -- & $\bullet$ & Healthline & -- & $\bullet$ \\
Booking & $\bullet$ & $\bullet$ & Huggingface & $\bullet$ & -- \\
CA.gov & -- & $\bullet$ & iclr & -- & $\bullet$ \\
Cambridge Dictionary & $\bullet$ & -- & IGN & -- & $\bullet$ \\
CarMax & -- & $\bullet$ & IKEA & -- & $\bullet$ \\
Carnival Cruise & -- & $\bullet$ & IMDb & -- & $\bullet$ \\
Cars.com & -- & $\bullet$ & Imgur & -- & $\bullet$ \\
Carvana & -- & $\bullet$ & Instructure & -- & $\bullet$ \\
Cboe & -- & $\bullet$ & IRS Refund Tracker & -- & $\bullet$ \\
CBP.gov & -- & $\bullet$ & JCPenney & -- & $\bullet$ \\
Chase & -- & $\bullet$ & Kaggle & -- & $\bullet$ \\
Chess.com & -- & $\bullet$ & Kelley Blue Book & -- & $\bullet$ \\
Chronicle Jobs & -- & $\bullet$ & LandWatch & -- & $\bullet$ \\
CoinMarketCap & -- & $\bullet$ & League of Legends & -- & $\bullet$ \\
Compass & -- & $\bullet$ & Macy's Wine Shop & -- & $\bullet$ \\
Cookpad & -- & $\bullet$ &  &  &  \\
\bottomrule
\end{tabular}
\caption{Provenance of the 153 \ours environments (1/2, continued in Table~\ref{tab:full-envs-list-2}). A bullet under WV or OM2W marks that the target site appears in the corresponding WebVoyager or Online-Mind2Web source task set. Memberships are non-exclusive: the suite covers 15 WebVoyager and 147 Online-Mind2Web sites.}
\label{tab:full-envs-list}
\end{table}

\begin{table}[p]
\centering
\footnotesize
\setlength{\tabcolsep}{3pt}
\begin{tabular}{@{}>{\raggedright\arraybackslash}p{0.35\linewidth}cc@{\quad}>{\raggedright\arraybackslash}p{0.35\linewidth}cc@{}}
\toprule
Environment & WV & OM2W & Environment & WV & OM2W \\
\midrule
Marriott & -- & $\bullet$ & Statista & -- & $\bullet$ \\
Mayo Clinic & -- & $\bullet$ & Steam & -- & $\bullet$ \\
Medicare.gov & -- & $\bullet$ & StubHub & -- & $\bullet$ \\
MEGA & -- & $\bullet$ & Student.com & -- & $\bullet$ \\
Megabus & -- & $\bullet$ & Super Lawyers & -- & $\bullet$ \\
Merriam-Webster & -- & $\bullet$ & Target & -- & $\bullet$ \\
Michaels & -- & $\bullet$ & TED & -- & $\bullet$ \\
Micro Center & -- & $\bullet$ & The Weather Network & -- & $\bullet$ \\
Microsoft Azure & -- & $\bullet$ & Thumbtack & -- & $\bullet$ \\
MTA & -- & $\bullet$ & Ticketmaster & -- & $\bullet$ \\
NBA & -- & $\bullet$ & TourRadar & -- & $\bullet$ \\
NFL & -- & $\bullet$ & Trader Joe's & -- & $\bullet$ \\
NVIDIA & -- & $\bullet$ & Trip.com & -- & $\bullet$ \\
NYSE & -- & $\bullet$ & Tumblr & -- & $\bullet$ \\
Ohio State University & -- & $\bullet$ & U.S. DOJ & -- & $\bullet$ \\
Ohio.gov & -- & $\bullet$ & UC Berkeley & -- & $\bullet$ \\
OhioMeansJobs & -- & $\bullet$ & Uniqlo & -- & $\bullet$ \\
Parkers & -- & $\bullet$ & United Airlines & -- & $\bullet$ \\
Petfinder & -- & $\bullet$ & University of Michigan & -- & $\bullet$ \\
PhET Interactive Simulations & -- & $\bullet$ & UPS & -- & $\bullet$ \\
Phys.org & -- & $\bullet$ & US Appliance & -- & $\bullet$ \\
Porsche & -- & $\bullet$ & USCIS.gov & -- & $\bullet$ \\
Public Storage & -- & $\bullet$ & USPS.com & -- & $\bullet$ \\
Qatar Airways & -- & $\bullet$ & Verizon & -- & $\bullet$ \\
Raising Cane's & -- & $\bullet$ & Versus & -- & $\bullet$ \\
Recreation.gov & -- & $\bullet$ & Virginia DMV & -- & $\bullet$ \\
Red Bull & -- & $\bullet$ & Walmart Careers & -- & $\bullet$ \\
REMAX & -- & $\bullet$ & Wanderlog & -- & $\bullet$ \\
RottenTomatoes & -- & $\bullet$ & Weather & -- & $\bullet$ \\
Ryanair & -- & $\bullet$ & WebMD & -- & $\bullet$ \\
Samsung & -- & $\bullet$ & WebMD Doctor & -- & $\bullet$ \\
SEC.gov & -- & $\bullet$ & Wine Access & -- & $\bullet$ \\
SmartAsset & -- & $\bullet$ & Wolfram Alpha & $\bullet$ & -- \\
SoundCloud & -- & $\bullet$ & Y Combinator & -- & $\bullet$ \\
SourceForge & -- & $\bullet$ & Yahoo Finance & -- & $\bullet$ \\
Speedo & -- & $\bullet$ & YouTube & -- & $\bullet$ \\
SpotHero & -- & $\bullet$ & Zara & -- & $\bullet$ \\
Stanford University & -- & $\bullet$ & ZipRecruiter & -- & $\bullet$ \\
\bottomrule
\end{tabular}
\caption{Provenance of the 153 \ours environments (2/2, continued from Table~\ref{tab:full-envs-list}).}
\label{tab:full-envs-list-2}
\end{table}

%% file: tables/rl_browser_action_space.tex
\begin{table}[h!]
\centering
\footnotesize
\begin{tabularx}{\linewidth}{@{}l>{\raggedright\arraybackslash}X@{}}
\toprule
Tool & Arguments and function \\
\midrule
\multicolumn{2}{@{}l}{\textit{Pointer}} \\
\texttt{click} & \texttt{point\_2d, clicks, button}. Click a viewport point with the chosen mouse button. \\
\texttt{hover} & \texttt{point\_2d}. Move the pointer to reveal hover states, tooltips, or menus. \\
\texttt{drag} & \texttt{start\_point\_2d, end\_point\_2d, steps}. Drag between two viewport coordinates. \\
\midrule
\multicolumn{2}{@{}l}{\textit{Keyboard}} \\
\texttt{write} & \texttt{message}. Clear the focused field and enter text. \\
\texttt{press\_keys} & \texttt{keys, is\_hotkey, presses}. Execute sequential keys or a keyboard shortcut. \\
\midrule
\multicolumn{2}{@{}l}{\textit{Navigation}} \\
\texttt{scroll} & \texttt{direction, amount, point\_2d}. Scroll the page or an element by a viewport fraction. \\
\texttt{goto\_url} & \texttt{url}. Navigate the active tab directly to a URL. \\
\texttt{go\_back} & No arguments. Return to the previous browser history entry. \\
\texttt{wait} & \texttt{seconds}. Pause for page loading or animation. \\
\midrule
\multicolumn{2}{@{}l}{\textit{Tabs}} \\
\texttt{new\_tab} & No arguments. Open a blank browser tab. \\
\texttt{switch\_tab} & \texttt{tab\_index}. Activate a tab by its zero-based index. \\
\texttt{close\_tab} & No arguments. Close the active tab and select a remaining tab. \\
\midrule
\multicolumn{2}{@{}l}{\textit{Termination}} \\
\texttt{done} & \texttt{response}. End the episode and return the final user-facing answer. \\
\bottomrule
\end{tabularx}
\caption{Browser action space shared by live-web and \ours mirror rollouts.}
\label{tab:rl-browser-action-space}
\end{table}

%% file: tables/rl_training_hyperparameters.tex
\begin{table}[t]
\centering
\footnotesize
\begin{tabularx}{\linewidth}{@{}>{\raggedright\arraybackslash}p{0.30\linewidth}>{\raggedright\arraybackslash}X@{}}
\toprule
Parameter & \ours RL setting \\
\midrule
\multicolumn{2}{@{}l}{\textit{Initialization and environment}} \\
Initialization & \texttt{OpenWebRL/OpenWebRL-4B-SFT} \\
Training set & 2,288 tasks from 123 \ours websites, with 15 WebVoyager and 15 other websites held out \\
Episode environment & One isolated \ours mirror sandbox initialized from a frozen seed \\
Viewport & $1280\times1000$ pixels, device-pixel ratio 1 \\
\midrule
\multicolumn{2}{@{}l}{\textit{Rollout and context}} \\
RL iterations & 100 \\
Effective task groups per iteration & 48 groups retained after dynamic sampling \\
Trajectories per task & 5 \\
Maximum browser steps & 50 \\
Screenshots in policy context & Current screenshot only \\
History in policy context & Full prior reasoning, tool calls, and textual environment feedback \\
Maximum context length & 32,768 tokens \\
Maximum response length & 1,024 tokens per policy turn \\
Rollout sampling & Temperature 0.8, top-$p$ 1.0, top-$k$ disabled \\
\midrule
\multicolumn{2}{@{}l}{\textit{MM-GRPO optimization}} \\
Advantage estimator & Group-relative trajectory-reward advantage, applied to all assistant turns \\
Dynamic sampling & Discard task groups with zero reward variance \\
PPO epochs & 2 \\
Asymmetric clipping & $\epsilon_{\mathrm{low}}=0.20$, $\epsilon_{\mathrm{high}}=0.28$ \\
KL / entropy coefficients & 0 / 0 \\
Global optimization batch size & 256 assistant-turn samples \\
Optimizer & Adam, $\beta=(0.9,0.98)$, weight decay 0.1 \\
Learning rate & $1\times10^{-6}$ with a constant schedule \\
Training and rollout backends & Megatron (TP2, DP4) and eight single-GPU SGLang engines \\
\midrule
\multicolumn{2}{@{}l}{\textit{Reward}} \\
Optimized reward & OpenWebRL format gate followed by a binary GPT-4.1 success judgment \\
Judge evidence & Final answer, complete action history, and three recent screenshots \\
Task-specific \ours verifier & Logged as a diagnostic metric only and excluded from the policy objective \\
\bottomrule
\end{tabularx}
\caption{RL training configuration of \ours-4B, reusing the OpenWebRL MM-GRPO recipe and browser interface.}
\label{tab:rl-training-hyperparameters}
\end{table}